\documentclass[sigconf]{acmart}
\AtBeginDocument{%
  \providecommand\BibTeX{{%
    \normalfont B\kern-0.5em{\scshape i\kern-0.25em b}\kern-0.8em\TeX}}}

\usepackage{algorithm}
\usepackage{algorithmic}

\usepackage{booktabs}       % professional-quality tables
\usepackage{amsfonts}       % blackboard math symbols
\usepackage{nicefrac}       % compact symbols for 1/2, etc.
\usepackage{microtype}      % microtypography
\usepackage{amsmath}
\usepackage{graphicx}
\usepackage[ruled,vlined,algo2e,linesnumbered]{algorithm2e}
\usepackage{xcolor}
\usepackage{bm}
\usepackage{booktabs} % for professional tables
\usepackage{multirow}
\usepackage{makecell}

\usepackage{amssymb}

\def\etal{\emph{et al.}}

\def\prox{\text{Prox}}
\def\1{{\bf{1}}}
\def\0{{\bf{0}}}
\def\I{{\bf{I}}}

\def\u{{\bf u}}
\def\v{{\bf v}}

\def\s{{\bf s}}

\def\cS{\mathcal{S}}

\def\cW{{\mathcal{W}}}
\def\cL{{\mathcal{L}}}

\DeclareMathOperator*{\argmin}{arg\,min}

\usepackage{subcaption}

\copyrightyear{2023}
\acmYear{2023}
\setcopyright{acmlicensed}
\acmConference[MM '23] {Proceedings of the 31st ACM International Conference on Multimedia}{October 29--November 3, 2023}{Ottawa, ON, Canada.}
\acmBooktitle{Proceedings of the 31st ACM International Conference on Multimedia (MM '23), October 29--November 3, 2023, Ottawa, ON, Canada}
\acmPrice{15.00}
\acmISBN{979-8-4007-0108-5/23/10}
\acmDOI{10.1145/3581783.3611838}
\acmSubmissionID{65}

\begin{document}

\title[Minimax Optimization for SNN Compression]{Resource Constrained Model Compression via Minimax Optimization for Spiking Neural Networks}

\author{Jue Chen}
\affiliation{
\institution{Peking University 
% \\ Kuaishou Technology
}
  \city{Beijing}
  \country{China}
  }
\email{chenjue@stu.pku.edu.cn}

% \author{Huan Yuan, Jianchao Tan$^*$, \\Bin Chen, Chengru Song, Di Zhang}
% \affiliation{
% \institution{Kuaishou Technology}
%   \city{Beijing}
%   \country{China}
%   }
% \email{{yuanhuan, jianchaotan}@kuaishou.com}

\author{Huan Yuan}
\affiliation{
\institution{Kuaishou Technology}
  \city{Beijing}
  \country{China}
  }
\email{yuanhuan@kuaishou.com}

\author{Jianchao Tan}
\authornote{The corresponding author.}
\affiliation{
\institution{Kuaishou Technology}
  \city{Beijing}
  \country{China}
  }
\email{jianchaotan@kuaishou.com}

\author{Bin Chen}
\affiliation{
\institution{Kuaishou Technology}
  \city{Beijing}
  \country{China}
  }
\email{chenbin08@kuaishou.com}

\author{Chengru Song}
\affiliation{
\institution{Kuaishou Technology}
  \city{Beijing}
  \country{China}
  }
\email{songchengru@kuaishou.com}

\author{Di Zhang}
\affiliation{
\institution{Kuaishou Technology}
  \city{Beijing}
  \country{China}
  }
\email{zhangdi08@kuaishou.com}


\begin{abstract}
    Brain-inspired Spiking Neural Networks (SNNs) have the characteristics of event-driven and high energy-efficient, which are different from traditional Artificial Neural Networks (ANNs) when deployed on edge devices such as neuromorphic chips. Most previous work focuses on SNNs training strategies to improve model performance and brings larger and deeper network architectures. It's difficult to deploy these complex networks on resource-limited edge devices directly. To meet such demand, people compress SNNs very cautiously to balance the performance and the computation efficiency. Existing compression methods either iteratively pruned SNNs using weights norm magnitude or formulated the problem as a sparse learning optimization. We propose an improved end-to-end Minimax optimization method for this sparse learning problem to better balance the model performance and the computation efficiency. We also demonstrate that jointly applying compression and finetuning on SNNs is better than sequentially, especially for extreme compression ratios. The compressed SNN models achieved state-of-the-art (SOTA) performance on various benchmark datasets and architectures. Our code is available at \textit{\url{https://github.com/chenjallen/Resource-Constrained-Compression-on-SNN}}.
\end{abstract}





\maketitle

\section{Introduction}
As the third generation of neural networks~\cite{thirdgeneration}, Spiking Neural Networks (SNNs) has been widely concerned in recent years. SNNs are the core of brain heuristic intelligence research, which have high biological interpretability and strong Spatio-temporal information processing ability~\cite{recentadvances}. In addition, due to the inherent asynchrony and sparsity of spiking training, these types of networks can maintain relatively good performance as well as low power consumption, especially when combined with neuromorphic chips~\cite{lowpowerconsumption, efficientneuromorphic}. With the development of efficient deep SNN training strategies, some useful network architectures are built, such as Spiking ResNet~\cite{spikingresnethu, spikingresnetzheng, spikingresnethulee} and SEW ResNet~\cite{sewresnet} to improve the performance of SNNs. The parameters and computational energy of SNN models rapidly increase, while the computational resources of edge devices are usually limited. For example, SpiNNaker demonstrated to run networks with up to 250,000 neurons and 80 million synapses on a 48-chip board~\cite{spinnaker}, which is still unable to run those more advanced SNNs. Thus, it is of great significance to compress SNNs before deploying them in real scenarios, which reduces computing costs, saves storage resources, and helps researchers exploit more benefits from high energy savings.
% model compression
Model compression was proposed to reduce the model size and improve the inference efficiency of the DNNs~\cite{han2015deep}.
%Because of the effectiveness, 
Weights pruning~\cite{han2015learning} is one of the widely used techniques for compressing the model size by zeroing out the individual weight of the convolutional kernel or fully connected weights matrix.
%Weights pruning~\cite{han2015learning, han2015deep} is effective to reduce the model size by zeroing out the individual weight of the convolutional kernel or fully connected weights matrix. Such non-structural sparsity does not lead to inference speedup on general implementations and hardware.
Filter pruning~\cite{li2016pruning, liu2017learning, yang2018netadapt} is another kind of pruning technique that prunes entire filters (or nodes for fully connected layers) and their corresponding weights. In this way, the entire filters can be removed and the original DNN can be transformed to be a thinner network, thus achieving speedup on general hardware.

% snn pruning.
Recently, researchers have carried out several works on SNNs pruning methods and made considerable progress. 
In GPSNN~\cite{gpsnn2015}, a two-stage growing-pruning algorithm was used to compress fully connected SNN so that the network could achieve better performance. 
In~\cite{stdpbased2019}, the non-critical synapses of SNNs were regularly pruned during Spike Timing Dependent Plasticity (STDP) training process based on a preset threshold. 
A soft pruning method has been considered to reduce the number of SNN weight updating during network training~\cite{softpruning2019}. 
Recently, ADMM optimization combined with Spatio-temporal backpropagation (STBP) training was used to compress SNNs~\cite{admm2021}. 
An attention-guided compression technique presented in~\cite{spikethrift2021}, used two steps to generate compressed deep SNN that could reduce spiking activity. 
Recent work~\cite{spatiotemporalpruning2021} performs pruning on the temporal dimension of SNNs to reduce time steps for inference. 
Grad Rewiring~\cite{gradr2021} is a joint learning algorithm of SNN connection and weight, which can strengthen the exploration of network structures. 
Most existing SNNs pruning work has either focused on shallow structures or has only attempted to prune networks at low sparsity. 
Besides, A very recent work proposed a dynamic pruning framework to prune SNNs based on temporal lottery ticket hyperthesis~\cite{dynsnn2022}, which handles the weights pruning of the deep SNN structures. 

\begin{figure}[t]
\centering
\includegraphics[width=\columnwidth]{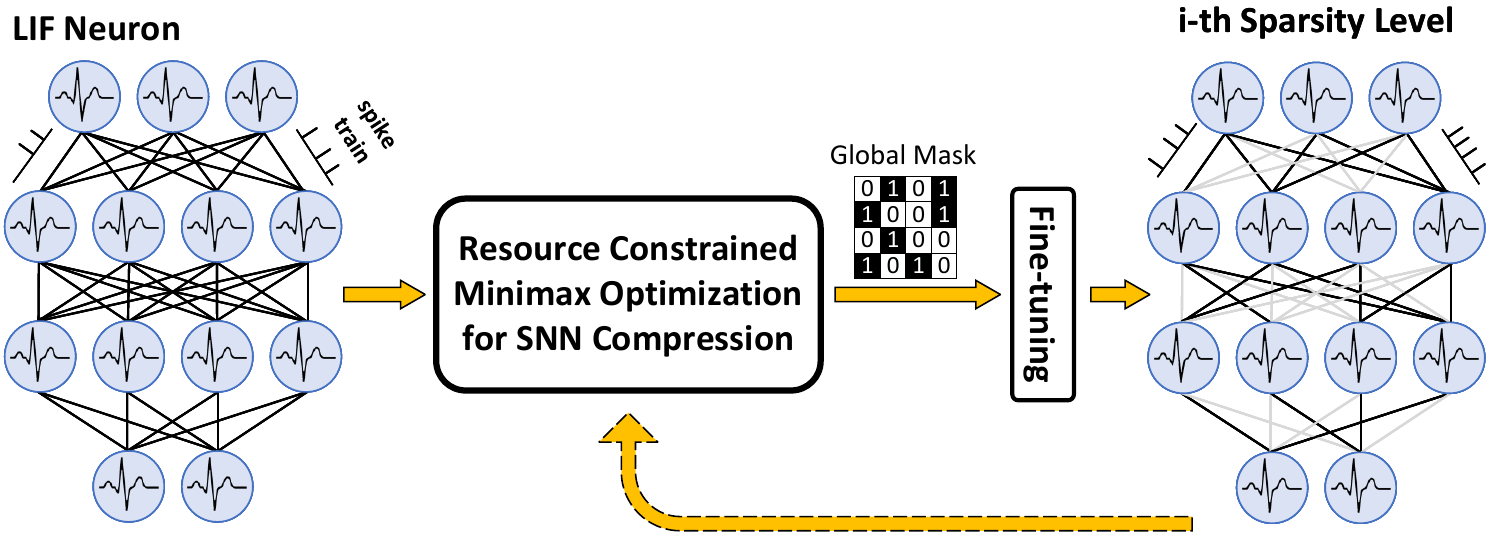}
\caption{Our whole pipeline. The resource-constrained Minimax Optimization method can compress SNNs into a lightweight model with different sparsity levels.}
\label{fig:teaser}
\end{figure}

% our method
In this paper, we present an end-to-end framework of weights pruning 
% \rev{(and filter pruning in the supplementary materials)} 
to compress the SNNs with a given resource budget. Unlike most resource-constrained compression methods which treat the resource consumption function as a black box~\cite{he2018amc}, we directly use the resource consumption to formulate a constrained optimization. The key idea is to use learnable parameters to control the lower bound of the sparsities. This introduces a sparsity constraint so that the resource constraint will only depend on the sparsity parameters. The constrained problem can be transformed into a Minimax optimization. Since the sparsity and resource constraints are not differentiable, the Minimax problem cannot be directly solved by gradient-based methods. In this work, we use the difference of convex function (DC)~\cite{tono2017efficient} sparsity reformulation and straight-through estimator (STE)~\cite{bengio2013estimating} to build a gradient-based algorithm to effectively optimize the compression problem.

We summarize the contributions as below:
\begin{itemize}
\item We propose an end-to-end Minimax optimization method to successfully compress the SNNs, as shown in Figure~\ref{fig:teaser}. DC sparsity reformulation~\cite{tono2017efficient} and STE~\cite{bengio2013estimating} are key components in this Minimax reformulation. Our compression procedure is end-to-end joint training of compression and fine-tuning on SNNs.
\item We formulate the resource-constrained SNNs compression problem into a constrained optimization problem where the SNNs weights and resource consumption are linked with learnable sparsity parameters.
\item The algorithm is gradient-based and easy to train. Evaluations of SNNs pruning on the public benchmark tasks show that our method is effective to compress SNNs and achieves state-of-the-art (SOTA) performance. 
%And our method can be generalized to be applied to other vision tasks.
\end{itemize}

\section{Related Work}
We review the related work from three aspects: the set of work in SNNs; the set of work in model compression and the specific model compression for SNNs. 

\subsection{Spiking Neural Networks}
Different from ANNs, SNNs have a temporal dimension inherently, which uses sparse binary spike event sequences to represent information. Therefore, they contribute to more energy savings in specialized neuromorphic hardware~\cite{efficientneuromorphic}. The information is transmitted among neurons via synapses. When the membrane potential exceeds a certain threshold caused by accumulating received spikes, the neuron fires a spike to the next layer. In this study, we employed the Leaky Integrate-and-Fire (LIF) neuron~\cite{lifmodel}, which is one of the most widely used neurons due to its effectiveness. The most common form of the LIF neuron is described as:
\begin{align}
\tau_{m} \frac{\mathrm{d} V_{m}(t)}{\mathrm{d} t}=-\left(V_{m}(t)-V_{\mathrm{rest}}\right) + X_{t}
\end{align}
where $V_{m}(t)$ represents the membrane potential of the neuron at time $t$, $X_{t}$ represents the input from the presynaptic neuron. $\tau_{m}$  is the membrane time as a constant value, that controls the decay and $V_{\mathrm{rest}}$ is the resting potential after firing. A spike will fire if $V_{t}$ exceeds the threshold $V_{\mathrm{th}}$. As claimed in previous works~\cite{wutrainfast} and~\cite{gradr2021}, We convert the above continuous differential equation into a discrete version:
\begin{align} 
H_{t+1} &=V_{t}+\frac{1}{\tau_{m}}\left(-\left(V_{t}-V_{\mathrm{rest}}\right)+ X_{t}\right) 
\\ S_{t+1} &=\Theta\left(m_{t+1}-V_{\mathrm{th}}\right) 
\\ V_{t+1} &=S_{t+1} V_{\mathrm{rest}}+\left(1-S_{t+1}\right) H_{t+1} 
\end{align}
where $H_{t}$ and $V_{t}$ denote the value of membrane potential after neural dynamics and after generating a spike at time step $t$, respectively. $S_{t}$ denotes the spike output at time step $t$. $\Theta(\cdot)$ is the Heaviside step function which is defined as $\Theta(x)=1$ for $x >= 0$ and $\Theta(x)=0$ for $x < 0$.

As we can see, the integration and firing behavior of neurons will result in the non-differentiability of the transfer function. So it is difficult to apply standard backpropagation in the training phase~\cite{BOHTE200217}. To obtain a high-performance SNN, researchers have proposed various training methods~\cite{diehlann2snn2015,ann2snncao2015,ann2snnhu2021,ann2snnli2021,stdpMasquelier2007,bplee2016}. Recently, some works focus on supervised learning based on backpropagation algorithms, where they use a surrogate gradient function to approximate the gradient of non-differentiable spike activity~\cite{stbp,spikebasedbp,fang2021learnmem,li2021differentiable,tettraining,autosnn}. These surrogate gradient methods provide an effective solution for training SNNs with deeper architecture~\cite{deepersnn2020}, such as VGGNet~\cite{vgg} and ResNet~\cite{resnet} families. Therefore, we adopt a backpropagation algorithm based on surrogate gradient~\cite{stbp} as the basic method for our SNNs training.

\begin{figure}[t]
    \centering
    \includegraphics[width=1.0\columnwidth]{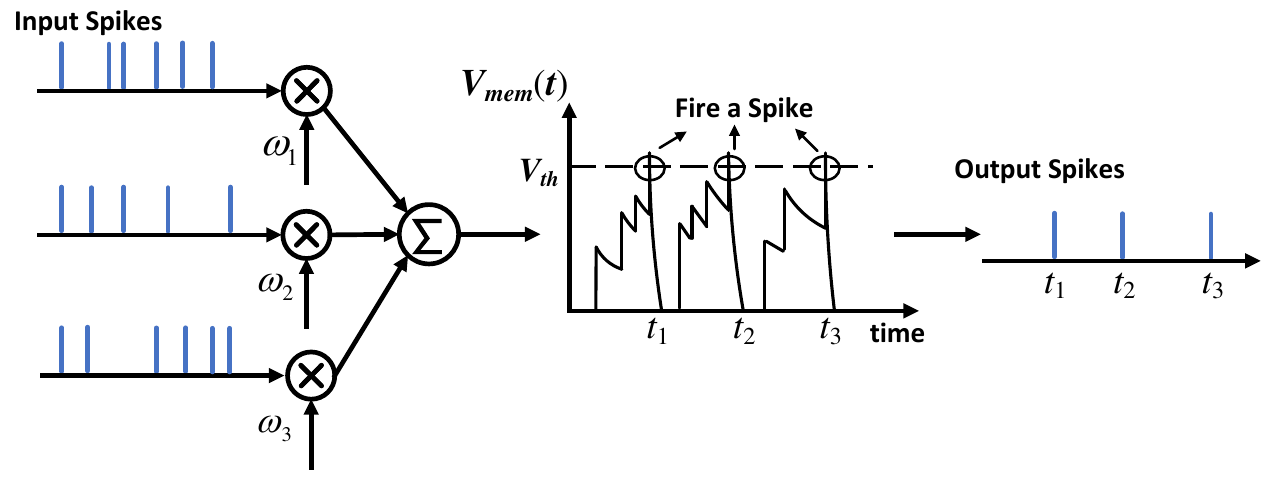} 
    \caption{The dynamics of LIF neurons, as similarly described in~\cite{lifmodelfigure}. When the membrane potential exceeds a threshold value, the neuron will fire a spike to the next layer and resets.}
    \label{fig:snn_neuron}
\end{figure}

\subsection{Model Compression}
%Such non-structural sparsity does not guarantee to speed up inference time on general implementations and hardware.
% Some works combine channel pruning and architecture search ~\cite{he2017channel} or ``neuron pruning''~\cite{zhou2016less} in their papers, but they usually refer to similar ideas. Wen~\etal~\cite{wen2016learning} investigate more on structures to accelerate the convolution operation and use $\ell_1$-norm based regularization to prune them.
% skip configuration~\cite{dong2021attention,yu2021unified}, etc.
There are different techniques for model compression, such as pruning~\cite{han2015learning, zhou2016less, he2017channel,miao2021learning, li2022revisiting, zhang2022carrying, hou2022chex}, quantization~\cite{han2015deep, hubara2017quantized,cai2019once}, low-rank factorization~\cite{lebedev2014speeding, li2019learning}, etc. Pruning utilizes the sparsity of weights tensors to achieve model compression. Weights pruning~\cite{han2015learning, xiao2019autoprune} is effective to remove the single elements by zeroing out the individual weight. Moreover, structured pruning~\cite{li2016pruning,guo2021gdp,Ding_2021_ICCV,huang2021rethinking, yu2021auto} prunes the weights according to specific structures to achieve more speedup on general hardware. Filter pruning~\cite{molchanov2016pruning, zhuang2018discrimination, liu2017learning, ye2018rethinking} is the most broadly used structured pruning, which prunes all the weights associated with certain filters.  Most filter pruning works prune the channels with low magnitude weights~\cite{IMP2018lottery, he2019filter}, or estimate the importance of channels for pruning~\cite{yu2020autoslim, molchanov2019importance, liu2021group}. Yang~\etal~\cite{yang2018netadapt} pre-compute a lookup table to estimate the latency of a convolution layer with different input/output channels, and use a greedy search strategy to gradually reduce the number of filters until the given resource budget is achieved. He~\etal~\cite{he2018amc} adopt reinforcement learning to search the number of pruned filters for each layer. The classification accuracy is used as the reward, and the number of pruned filters is taken as the action. Recently, more approaches~\cite{yang2018netadapt,he2018amc,liu2018rethinking} consider the model compression as a constrained problem~\cite{yang2019ecc, shen2020umec,guo2021gdp,yu2021unified}. Furthermore, resource consumption is used to restrict the action space. These methods are successfully applied to fully supervised tasks such as image classification and object detection. 
In this paper, we proposed an end-to-end optimization method to solve a resource-constrained compression problem on SNNs, we demonstrate the problem formulation from the unstructured weights pruning perspective.

\subsection{Model Compression for SNN}
%Lottery tickets, ADMM, Gradient Rewiring, and so on
To reduce the energy consumption of SNNs, some approaches focus on the compression of SNN models recently, such as connection pruning~\cite{admm2021, gradr2021, kim2022lottery} and model quantization~\cite{quantsnn2020exploring, quantsnn2020optimizing, admm2021, xu2022ultra}. Deng~\etal~\cite{admm2021} defined the connection pruning and weight quantization as a constrained optimization problem and used Spatio-temporal backpropagation (STBP) and alternating direction method of multipliers (ADMM) to solve it. Chen~\etal~\cite{gradr2021} formulated the gradient rewiring (Grad R) algorithm which redefines the gradient of SNNs to a new synaptic parameter and joint learning SNNs connection and weight. In addition, Kim~\etal~\cite{kim2022lottery} performed the connection pruning toward deeper SNNs ( $\geq$ 16 layers) and combined the Iterative Magnitude Pruning~\cite{IMP2018lottery} and Early-Bird~\cite{EB2019drawing} tickets to obtain smaller SNNs.
Chen~\etal~\cite{stdschen22} proposed a dynamic pruning algorithm based on nonlinear reparameterization mapping from spine size to SNN weights.
To compare with these unstructured weights pruning works for SNNs, we adopt the Minimax optimization method to jointly optimize the global sparsity of SNNs and weights parameters. We handle all layers' weights parameters globally with one sparsity parameter to solve the unstructured pruning problem.

\section{Formulation}

\subsection{Resource-Constrained Optimization}
The ideal scenario of SNN compression is that given a resource budget $R_{\text{budget}}$ (based on a certain metric, e.g., Parameters, Flops, latency, or energy), the compression method can return a compressed model which satisfies the given budget and maintains the accuracy as well as possible. The whole process should be automatic, i.e., there is no need to manually set the sparsity of each layer. In this paper, we directly formulate such a compression scheme for a constrained optimization problem:
\begin{subequations}
\label{eq:obj1}
\begin{align}
\min_{\cW, \s}&\quad \cL(\cW ) \label{eq:obj1_min}\\
\text{s. t.}& \quad R(s) \leq R_{\text{budget}}, \label{eq:obj1_con1}\\
& \sum_i \I({\cW_{i}} = 0) \geq s
\label{eq:obj1_con2}
\end{align}
\end{subequations}
where $R(s)$ evaluates a general resource consumption (e.g., Flops or latency) based on the number of (nonzero) weights for each layer. It does not need to be differentiable. For example, when representing latency, it can be computed by a latency table as in~\citet{yang2018netadapt}. $\cL$ is the standard training loss.

$\I(\cdot)$ is the indicator function that returns 1 if the argument is satisfied and 0 otherwise.
 $\s$ is a learnable scalar variable to control the lower bound of the sparsity of weight parameters vector $\cW$ of the whole network. This formulation holds because the resource function $R$ monotonically decreases with respect to the increasing sparsities,
%This is usually true for the filter pruning case,
i.e., the more weights are pruned, the smaller the resource consumption we have.
Note that we mainly focus on the unstructured pruning (or called weights sparsification) for SNN in the main text, thus $\cW_i$ in the above equation stands for \textbf{each element of weight parameters vector $\cW$} of the whole network. 
% \rev{We will show comparisons to the results of structured channel pruning in the supplementary materials.}

\section{Optimization}
In the previous section, we have already formulated the resource-constrained pruning as a constrained optimization~\eqref{eq:obj1}. In this section, we first do some reformulation to make it more convenient to solve. Then we propose a gradient-based algorithm to solve the resource-constrained pruning in an end-to-end way.
\subsection{Minimax Reformulation}
The sparsity constraint~\eqref{eq:obj1_con2} is non-convex and the non-continuous indicator function makes it more difficult. Common approaches to deal with this constraint include $\ell_1$-norm relaxation~\cite{tibshirani1996regression} and $\ell_0$-norm projection~\cite{yuan2014gradient}. The ideas of $\ell_1$-norm relaxation have been applied in DNN compression~\cite{liu2017learning, ye2018rethinking, huang2018data}. However, the $\ell_1$-norm can only approximate the sparsity constraint, so it is not equivalent to the sparsity constraint and there is no guarantee to bound the real sparsity by restricting the $\ell_1$-norm.

Let $\|\u\|_{s, 2}$ be the bottom-($s$, 2) ``norm'', which denotes the $\ell_2$-norm of the sub vector composed of bottom-$s$ elements in magnitude. Then we have an {\em equivalent} reformulation for the sparsity constraint~\eqref{eq:obj1_con2}:
\begin{equation}
\|\cW\|_{s, 2} = 0
%\sum_{\uparrow \lceil \s^{(l)} \rceil} {\bm{\gamma}^{(l)}}^2 = 0 
\Leftrightarrow \sum_i {\I(\cW_i = 0)} \geq s, \label{eq:dc_sparsity}
\end{equation}
%where $\|\u\|_{s, 2}^2$ is the bottom-($s$, 2) norm, denotes the $\ell_2$-norm of the sub vector composed of bottom-$s$ elements in magnitude.

%where $\sum_{\uparrow s} (\u)$ is the sum of the least $s$ elements (in magnitude) of $\u$, and $\u^2$ is the element-wise square operation, i.e. $(\u^2)_i = u_i^2$.
Equation~\eqref{eq:dc_sparsity} is proved by Tono~\etal~\cite{tono2017efficient}, where the left-hand side is called DC (difference of convex functions) representation of the $\ell_0$-constraint. By this transformation, the sparsity constraint becomes an equality constraint of a continuous function. Compared to the original $\ell_0$-norm constraint, it can be written as a ``soft'' constraint and avoid being stuck in the bad local optimum of the constraint set.

\begin{figure}
    \begin{center}
        \includegraphics[width=0.7\linewidth]{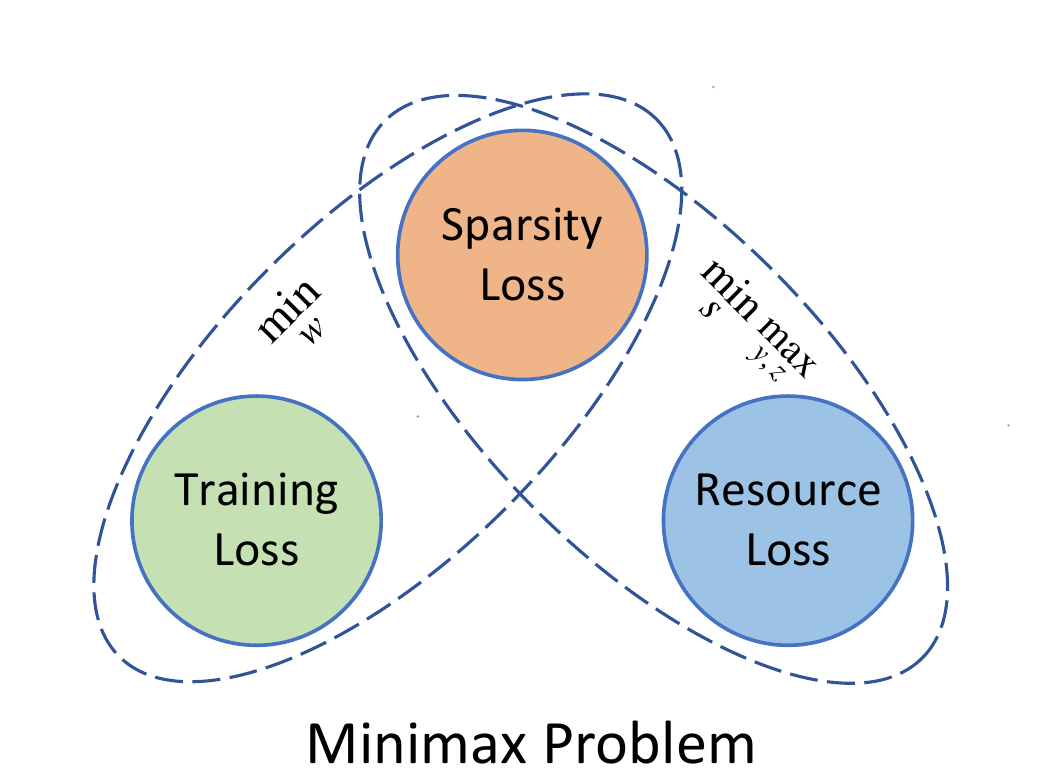}
    \end{center}
        \caption{The reformulated minimax problem. It consists of three loss terms which are training loss, sparsity loss, and resource loss. The sparsity loss and resource loss are used to restrict the SNN resource consumption.}
    \label{fig:fig1}
\end{figure}

By introducing dual variables $y$ and $z$, we derive the minimax reformulation of problem~\eqref{eq:obj1}:
% {\small
% \begin{equation}
% \min_{\cW, \bm{\gamma}, \s} \max_{\y, z\geq 0} \cL(\cW, \bm{\gamma}) + \underbrace{\sum_{l=1}^L \y^{(l)}\|\bm{\gamma}^{(l)}\|_{\lceil \s^{(l)} \rceil, 2}^2}_{\text{sparsity loss: }\cS(\y, \s, \bm{\gamma})} + \underbrace{z (R(\s) - R_{\text{budget}}).}_{\text{resource loss}} \label{eq:obj2}\\
% \end{equation}
% }
% {\small
% \begin{equation}
% \min_{\cW, \s} \max_{\y, z\geq 0} \cL(\cW) + \underbrace{\sum_{l=1}^L \y_{i}^{(l)}\|\bm{\cW}_{i,;}^{(l)}\|_{\lceil \s^{(l)} \rceil, 2}^2}_{\text{sparsity loss: }\cS(\y, \s)} + \underbrace{z (R(\s) - R_{\text{budget}}).}_{\text{resource loss}} \label{eq:obj2}\\
% \end{equation}
% }
{\small
\begin{equation}
\min_{\cW, s} \max_{y, z\geq 0} \cL(\cW) + \underbrace{y \|\cW\|_{\lceil s \rceil, 2}^2}_{\text{sparsity loss: }\cS(y, s, \cW)} + \underbrace{z (R(s) - R_{\text{budget}}).}_{\text{resource loss}} \label{eq:obj2}\\
\end{equation}
}
Where we introduce the sparsity loss $\cS(y, s, \cW):= y \|\cW\|_{\lceil s \rceil, 2}^2$, and resource loss $z (R(s) - R_{\text{budget}})$ to substitute the original constraints.  Figure~\ref{fig:fig1} shows an illustration of the reformulated minimax problem.
It is easy to verify that~\eqref{eq:obj2}$\rightarrow\infty$ if $\|\cW \|_{\lceil s \rceil, 2}^2 \neq 0$ or $R(s) - R_{\text{budget}} > 0$, and~\eqref{eq:obj2}$=$~\eqref{eq:obj1_min} if both constrains~\eqref{eq:obj1_con1} and~\eqref{eq:obj1_con2} are satisfied. With the fact~\eqref{eq:dc_sparsity}, we can see that~\eqref{eq:obj2} is an equivalent reformulation of the original problem~\eqref{eq:obj1}.

\subsection{Gradient-based Algorithm}
In deep learning, gradient-based algorithms (e.g. SGD) are widely used to optimize the DNNs. Gradient descent ascent (GDA), is widely used as a gradient-based method to solve minimax problems~\cite{jin2019minmax}. The basic idea is iteratively doing gradient descent on the minimization variables and gradient ascent on the maximization variables.

In problem~\eqref{eq:obj2}, the functions $\|\v\|_{s, 2}^2$ and $R(\s)$ are not differentiable to $s$ and $\s$, so the GDA method cannot be directly applied. Straight-through estimator (STE)~\cite{bengio2013estimating} is an effective workaround for using the gradient-based algorithm to optimize non-differentiable functions. The basic idea is using some simple proxy as the derivative of the non-differentiable part, thus the back-propagation can be used as in the differentiable objective. In our case, both $\|\v\|_{s, 2}^2$ and $R(\s)$ are not complicated although they are non-differentiable. For $\|\v\|_{s, 2}^2$, we use the numerical differentiation $\|\v\|_{s+1, 2}^2 - \|\v\|_{s, 2}^2$ as the proxy derivative of $\|\v\|_{s, 2}^2$ with respect to $s$:
\begin{equation}
{\tilde{\partial} \|\v\|_{s, 2}^2 \over \tilde{\partial} s} = \v_{\text{least-$\min(\text{Dim}(\v), s + 1)$}}^2 %\v_{\text{least-$\min(\text{Dim}(\v), \lceil s \rceil + 1)$}}^2,
\end{equation}
where $\v^2$ is the element-wise square of $\v$, $\text{Dim}(\v)$ is the dimensionality of $\v$, and $\v^2_{\text{least-$j$}}$ is the $j$-th least element in vector $\v^2$.

% For commonly used resource consumption function (e.g., Flops, model size), the non-differentiable part of $R$ is the ceiling function $\lceil s \rceil$, for which we use a common STE~\cite{bengio2013estimating}:
% \begin{equation}
% {\tilde{\partial} \lceil s \rceil \over \tilde{\partial} s} = 1
% \end{equation}

% For practical resource consumption like the inference latency, we follow the way of Yang~\etal~\cite{yang2018netadapt} to make a lookup table for estimating the latency of each layer. Specifically, we have a two-dimensional table $T^{(l)}$ for each layer $l$, and $T^{(l)}(i, j)$ stores the latency of layer $l$ with $i$ input channels and $j$ output channels. In this way, we can get the DNN latency estimation $R$ by summing up the latency of all the layers. The lookup table is non-differentiable, so we adopt its numerical differentiation to serve as the STE:
% \begin{align}
% &{\tilde{\partial} T(i, j) \over \tilde{\partial} i} = T(i,j)-T(i-1,j)\\
% &{\tilde{\partial} T(i, j) \over \tilde{\partial} j} = T(i,j)-T(i,j-1)
% \end{align}
% In our case, the number of input channels is $\c^{(l)}-\lceil \s^{(l-1)} \rceil$, and the number of output channels is $\d^{(l)}-\lceil \s^{(l)} \rceil$, where $\c^{(l)}$ and $\d^{(l)}$ are the original numbers of input and output channels. The latency consumption function is $R(\s)=\sum_{l=1}^LT^{(l)}(\c^{(l)}-\lceil \s^{(l-1)} \rceil, \d^{(l)}-\lceil \s^{(l)} \rceil)$

\begin{algorithm2e}[htbp]
% \small
\label{alg1}
 \SetAlgoLined
 \KwIn{Resource budget $R_{\text{budget}}$, learning rates $\eta_1, \eta_2, \eta_3, \eta_4$, number of total iterations $\tau$.}
 \KwResult{SNN weights $\cW^{*}$}
 Initialize $t=1$;\\
 Initialize $\cW^1$ (randomly or from a pre-trained model);\\
 %\While{True}{
  	%\tcp{Update DNN weights}
	\While{$t\leq\tau$}{
         %\While{$W$ has not converged}{
        $ \cW^{t+1} = \prox_{\eta_1 \cS(y^t, s^t, \cW)}(\cW^t - \eta_1 \hat{{\nabla}}_{\cW} \cL(\cW^t))$\tcp*{Proximal-SGD}
        $s^{t+1} = s^t - \eta_2 (\tilde{\nabla}_\s \cS(y^t, s^t, \cW^{t+1}) + \tilde{\nabla}_s z^{t}(R(s^{t}) - R_{\text{budget}}))$\tcp*{Gradient (STE) Descent}
        $y^{t+1} = {y}^{t} + \eta_3 \|{\cW}^{t+1}\|_{\lceil {s}^{t+1} \rceil, 2}^2$\tcp*{Gradient Ascent}
        $z^{t+1} = max(0, z^t + \eta_4 (R(s^{t+1}) - R_{\text{budget}}))$\tcp*{Gradient Ascent}
     }
 %}
$\cW^* = \cW$.
\caption{Gradient-based algorithm to solve~\eqref{eq:obj2}.}
\end{algorithm2e}

With these derivative estimators, we utilize a gradient-based algorithm to optimize~\eqref{eq:obj2}. The detailed steps are shown in Algorithm 1. We use proximal-SGD~\cite{nitanda2014stochastic} to update $\cW$. Specifically, first take gradient-descent to update $\cW$ to obtain $\bar{\cW}$, then the proximal operator $\prox_{\eta_1 \cS(y^t, s^t, \cW)}(\bar{\cW})$ for $\bar{\cW}$ is defined as:
\begin{equation}
\argmin_{{\cW}} {1\over 2} \| {\cW} - \bar{\cW}\|^2 + \eta_1 \cS(y^t, s^t, \cW)
\end{equation}
which has a closed-form solution $\cW'$:
\begin{equation}
\cW_i^{'} =
\begin{cases}
\bar{\cW}_i, \quad & \text{if $\bar{\cW}^{2}_i > \bar{\cW}^2_\text{least-$\lceil {s}^t \rceil$}$}
%$i$ is not one of the least $\lceil {\s^{(l)}}^t \rceil$ elements in $\bar{\bm{\gamma}}^{(l)}$};
\\
{1\over 1+2\eta_1 {y}^t}\ \bar{\cW}_i, &\text{otherwise}.
\end{cases}
\end{equation}

Our formulation gradually decays $\bar{\cW}_i$ by a factor ${1\over 1+2\eta_1 {y}^t}$, whose value is smaller than 1, instead of directly projecting the $\bar{\cW}_i$ into zero to meet with traditional $\ell_0$-constraint.
We use gradient descent (STE) to update $s$ and gradient ascent to update $y, z$, as described in the Algorithm~\ref{tab:first}.

\begin{table*}[!h]
    \caption{Performance comparison between our method and previous works on MNIST and CIFAR10 datasets.}
    \centering
    % \small
    \footnotesize
        \begin{tabular}{c c c c c c}
            \toprule
            \textbf{Pruning Method} & \textbf{Dataset} & \textbf{Arch.} & \textbf{Top-1 Acc. (\%)} & \textbf{Acc. Loss (\%)} &  \textbf{Sparsity (\%)} \\
            \midrule
                \multirow{3}{*}{ADMM-based~\cite{admm2021}} & 
                \multirow{3}{*}{MNIST} & 
                \multirow{3}{*}{LeNet-5 like}  & 
                \multirow{3}{*}{99.07} &  +0.03 & 50.00  \\ 
                & & & & -0.43 & 60.00  \\ 
                & & & & -2.23 & 75.00  \\
            \midrule
                \multirow{4}{*}{Deep R~\cite{deepr2018}} & 
                \multirow{4}{*}{MNIST} & 
                \multirow{4}{*}{2 FC}  & 
                \multirow{4}{*}{98.92} &  -0.36 & 62.86  \\ 
                & & & & -0.56 & 86.70  \\ 
                & & & & -2.47 & 95.82  \\
                & & & & -9.67 & 98.90  \\
            \midrule
                \multirow{5}{*}{Grad R~\cite{gradr2021}} & 
                \multirow{5}{*}{MNIST} & 
                \multirow{5}{*}{2 FC}  & 
                \multirow{5}{*}{98.92} & -0.33 & 74.29  \\ 
                & & & & -0.43 & 82.06  \\ 
                & & & & -2.02 & 94.37  \\
                & & & & -3.55 & 96.94  \\
                & & & & -8.08 & 98.62  \\
            \midrule
                \multirow{5}{*}{\textbf{Ours}} & 
                \multirow{5}{*}{MNIST} & 
                \multirow{5}{*}{2 FC}  & 
                \multirow{5}{*}{\textbf{98.91}} & \textbf{-0.06} & \textbf{75.00}  \\ 
                & & & & \textbf{-0.16} & \textbf{85.00}  \\ 
                & & & & \textbf{-1.23} & \textbf{95.00}  \\
                & & & & \textbf{-2.70} & \textbf{97.00}  \\
                & & & & \textbf{-7.34} & \textbf{98.70}  \\                    
            \midrule
            \midrule
                \multirow{3}{*}{ADMM-based~\cite{admm2021}} & 
                \multirow{3}{*}{CIFAR10} & 
                \multirow{3}{*}{7 Conv, 2 FC}  & 
                \multirow{3}{*}{89.53} &  -0.38 & 50.00  \\ 
                & & & & -2.16 & 75.00  \\ 
                & & & & -3.85 & 90.00  \\
            \midrule
                \multirow{3}{*}{Deep R~\cite{deepr2018}} & 
                \multirow{3}{*}{CIFAR10} & 
                \multirow{3}{*}{6 Conv, 2 FC}  & 
                \multirow{3}{*}{92.84} &  -1.98 & 94.76  \\ 
                & & & & -2.56 & 98.05  \\ 
                & & & & -3.53 & 98.96  \\
            \midrule
                \multirow{5}{*}{Grad R~\cite{gradr2021}} & 
                \multirow{5}{*}{CIFAR10} & 
                \multirow{5}{*}{6 Conv, 2 FC}  & 
                \multirow{5}{*}{92.84} & -0.30 & 71.59  \\ 
                & & & & -0.34 & 87.96  \\ 
                & & & & -0.81 & 94.92  \\
                & & & & -1.47 & 97.65  \\
                & & & & -3.52 & 99.27  \\
            \midrule
                \multirow{2}{*}{STDS~\cite{stdschen22}} & 
                \multirow{2}{*}{CIFAR10} & 
                \multirow{2}{*}{6 Conv, 2 FC}  & 
                \multirow{2}{*}{92.84} & -0.35 & 97.77 \\ 
                & & & & -2.63 & 99.25  \\ 
            \midrule
                \multirow{6}{*}{\textbf{Ours}} & 
                \multirow{6}{*}{CIFAR10} & 
                \multirow{6}{*}{6 Conv, 2 FC}  & 
                \multirow{6}{*}{\textbf{92.88}} & \textbf{+0.84} & \textbf{75.00}  \\ 
                & & & & \textbf{+0.52} & \textbf{88.04}  \\ 
                & & & & \textbf{+0.41} & \textbf{95.07}  \\
                & & & & \textbf{-0.13} & \textbf{97.71}  \\
                & & & & \textbf{-1.38} & \textbf{98.84}  \\
                & & & & \textbf{-2.56} & \textbf{99.31}  \\                    
            \bottomrule
            \end{tabular}
    \label{tab:first}
\end{table*}

\section{Experiments and Results}
%We evaluate our end-to-end SNNs compression method on four SNN models : MinstNet, Cifar10net, VGG16, and Resnet19.
We evaluate our end-to-end compression method for various SNN models on benchmark datasets. Our compressed SNN models achieved state-of-the-art (SOTA) performance against the previous best-performing SNNs compression methods in all cases. We put additional compression results and visualization plots in the Supplementary Materials.
% We compare our end-to-end Minimax optimization method against two baselines:

% \begin{itemize}
% \item Lasso regularization-based filter pruning. $\ell_1$-norm regularization (i.e. Lasso) of the normalization weights is used in many recently proposed filter pruning methods~\cite{liu2017learning, ye2018rethinking, huang2018data}. In the experiments, the original style transfer loss is added with a Lasso penalty term $\lambda \|\bm{\gamma}\|_1$, and we follow the ISTA method used in Ye~\etal~\cite{ye2018rethinking} to implement this baseline.
% \item Uniform filter pruning. A simple strategy to set the number of pruned filters is setting it proportionally to the number of original filters, i.e., $\sum_i \I(\gamma^{(l)}_i=0) = \alpha \d^{(l)}, \alpha \in (0,1)$. Recent filter pruning works~\cite{he2018soft, he2019filter} using such a strategy showed competitive performance. In our implementation, we prune the weights $\bm{\gamma}$ based on magnitude after setting the number of pruned filters for each layer.

% \end{itemize}
%$\ell_1$-norm regularization-based optimization and uniform channel pruning method. The Lasso regularization method is to directly add a penalty term on parameter sparsity. Uniform method prunes channels in a uniform sampling manner.
% We compare these four methods in terms of both style transfer results quality and their inference time on mobile devices: VIVO X9 mobile phone and \etc.

\begin{table*}[!h]
\caption{Performance comparison between our method and previous works on larger models.}
    \centering
    % \small
    \footnotesize
        \begin{tabular}{c c c c c c}
            \toprule
            \textbf{Pruning Method} & \textbf{Dataset} & \textbf{Arch.} & \textbf{Top-1 Acc. (\%)} & \textbf{Acc. (\%)} &  \textbf{Sparsity (\%)} \\
            \midrule
                \multirow{4}{*}{IMP~\cite{kim2022lottery}} & 
                \multirow{4}{*}{CIFAR10} & 
                \multirow{4}{*}{VGG16}  & 
                \multirow{4}{*}{-} &  92.66 & 68.30  \\ 
                & & & & 92.54 & 89.91  \\ 
                & & & & 92.38 & 95.69  \\
                & & & & 91.81 & 98.13  \\
            \midrule
                \multirow{5}{*}{\textbf{Ours}} & 
                \multirow{5}{*}{CIFAR10} & 
                \multirow{5}{*}{VGG16}  & 
                \multirow{4}{*}{\textbf{93.25}} & \textbf{93.27} & \textbf{69.02}  \\ 
                % & & & & \textbf{93.29} & \textbf{80.03}  \\ 
                & & & & \textbf{93.26} & \textbf{90.09}  \\
                & & & & \textbf{92.76} & \textbf{95.70}  \\
                & & & & \textbf{92.00} & \textbf{98.13}  \\                    
            \midrule
            \midrule
                \multirow{3}{*}{Grad R~\cite{gradr2021}} & 
                \multirow{3}{*}{CIFAR10} & 
                \multirow{3}{*}{ResNet19}  & 
                \multirow{3}{*}{93.22} &  92.68 & 76.90  \\ 
                & & & & 91.91 & 94.25  \\ 
                & & & & 91.12 & 97.56  \\
            \midrule
                \multirow{3}{*}{IMP~\cite{kim2022lottery}} & 
                \multirow{3}{*}{CIFAR10} & 
                \multirow{3}{*}{ResNet19}  & 
                \multirow{3}{*}{93.22} &  93.50 & 76.20  \\ 
                & & & & 93.46 & 94.29  \\ 
                & & & & 93.18 & 97.54  \\
            \midrule
                \multirow{3}{*}{\textbf{Ours}} & 
                \multirow{3}{*}{CIFAR10} & 
                \multirow{3}{*}{ResNet19}  & 
                \multirow{3}{*}{\textbf{94.85}} & \textbf{94.84} & \textbf{80.05}  \\ 
                & & & & \textbf{94.36} & \textbf{95.07}  \\ 
                & & & & \textbf{93.81} & \textbf{97.07}  \\                   
            \midrule
            \midrule
                \multirow{3}{*}{Grad R~\cite{gradr2021}} & 
                \multirow{3}{*}{CIFAR100} & 
                \multirow{3}{*}{ResNet19}  & 
                \multirow{3}{*}{71.34} &  69.36 & 77.03  \\ 
                & & & & 67.47 & 94.92  \\ 
                & & & & 67.31 & 97.65  \\
            \midrule
                \multirow{3}{*}{IMP~\cite{kim2022lottery}} & 
                \multirow{3}{*}{CIFAR100} & 
                \multirow{3}{*}{ResNet19}  & 
                \multirow{3}{*}{71.34} &  71.45 & 76.20  \\ 
                & & & & 71.00 & 94.29  \\ 
                & & & & 69.05 & 97.54  \\
            \midrule
                \multirow{3}{*}{\textbf{Ours}} & 
                \multirow{3}{*}{CIFAR100} & 
                \multirow{3}{*}{ResNet19}  & 
                \multirow{3}{*}{\textbf{74.71}} & \textbf{75.05} & \textbf{79.99}  \\ 
                & & & & \textbf{72.67} & \textbf{95.19}  \\ 
                & & & & \textbf{70.80} & \textbf{97.31}  \\      
            \midrule
            \midrule
                \multirow{2}{*}{Grad R~\cite{gradr2021}} & 
                \multirow{2}{*}{ImageNet} & 
                \multirow{2}{*}{SEW ResNet18}  & 
                \multirow{2}{*}{63.22} &  60.05 & 50.94  \\ 
                & & & & 24.62 & 53.65  \\ 
            \midrule
                \multirow{2}{*}{ADMM~\cite{admm2021}} & 
                \multirow{2}{*}{ImageNet} & 
                \multirow{2}{*}{SEW ResNet18}  & 
                \multirow{2}{*}{63.22} &  59.48 & 82.58  \\ 
                & & & & 55.85 & 88.84  \\ 
            \midrule
                \multirow{4}{*}{STDS~\cite{stdschen22}} & 
                \multirow{4}{*}{ImageNet} & 
                \multirow{4}{*}{SEW ResNet18}  & 
                \multirow{4}{*}{63.22} &  61.30 & 82.58  \\ 
                & & & & 59.93 & 88.84  \\ 
                & & & & 58.06 & 93.24  \\ 
                & & & & 56.28 & 95.30  \\  
            \midrule
                \multirow{4}{*}{\textbf{Ours}} & 
                \multirow{4}{*}{ImageNet} & 
                \multirow{4}{*}{SEW ResNet18}  & 
                \multirow{4}{*}{\textbf{63.25}} &  \textbf{61.42} & \textbf{82.50}  \\ 
                & & & & \textbf{60.51} & \textbf{88.84}  \\ 
                & & & & \textbf{58.12} & \textbf{93.20}  \\
                & & & & \textbf{56.46} & \textbf{94.39}  \\
            \bottomrule
            \end{tabular}
    \label{tab:second}
\end{table*}

\subsection{Implementation details}
In our works, we validate the compression method on six SNN models, including the shallow SNNs (e.g. 2 FC, 6 Conv and 2 FC) and the deep SNNs (e.g. VGG16~\cite{vgg}, ResNet19~\cite{resnet}, SEW ResNet18~\cite{sewresnet}, VGGSNN~\cite{tettraining}). 
We compare the performance of our method with previous SNN compression methods on static MNIST, CIFAR10, CIFAR100, ImageNet1K datasets and neuromorphic CIFAR10-DVS dataset which is converted from the static image dataset by using a DVS camera.
 Experiments are conducted on NVIDIA V100 GPUs and we use SpikingJelly~\cite{SpikingJelly} framework to implement SNNs. 

Similar to the previous SNN work~\cite{gradr2021}, we use a shallow SNN with 2 FC layers on the MINST dataset, and a model with 6 convolution layers and 2 FC layers for the CIFAR10 dataset. These two shallow SNNs are trained with Adam optimizer with a learning rate 1e-4. The timestep is set to 8. Other hyperparameters of baseline are the same as~\cite{gradr2021}(e.g. batch size, learning rate). What's more, we use deep SNNs, VGG16, ResNet19, and VGGSNN. The training method follows the previous SNN work~\cite{tettraining}. We train deep SNNs by SGD optimizer with momentum 0.9 and weight decay 5e-4. The learning rate is set to 0.05 for baseline training and cosine decay to 0. The timestep is set to 5 and the batch size is set to 32. The default number of epochs is set to 300.  
As for training SEW ResNet18 on ImageNet, we follow all the training setting of SEW ResNet~\cite{sewresnet}. The base learning is 0.1 with cosine annealing scheduler. The number of epochs is 320, the batch size is set to 32, and the timestep is set to 4.
% \rev{Considering the training cost of deep SNNs on ImageNet, we select SEW ResNet18 as a testbed and follow all the training setting of SEW ResNet.}
In all datasets, the learning rate of $y$ is set to 0.1, while the default learning rate of $z$ is set to $10^5$. We count the number of zero and nonzero values of the whole weights in SNN and compute the percentage of zero values to be the sparsity.

\paragraph{Training methodology}
According to previous SNNs work settings respectively, we first train SNNs models to get pre-trained baseline models. Then our compression training stage starts with pre-trained models. Before pruning the SNN models, we set a budget list which values are some compression ratios from 0.5 to 0.005 (e.g. [0.25, 0.1, 0.05, 0.01, 0.005]). Furthermore, the value of the budget list can be connectivity, parameter size, latency times, FLOPs, and so on. For connection pruning, the values in the budget list are connectivity ratios. For structure pruning, the values in the budget list are FLOPs ratios in our work. 
During our compression training, if the model connectivity meets the current target compression ratio, we pause the pruning and then fine-tune the snapshot at this ratio until achieving the maximum fine-tuning epochs.
After the fine-tuning process at the current compression rate finishes, our method removes the current ratio from the budget list and then continues the compression training automatically to achieve the next compression ratio. Compression and fine-tuning are jointly performed in the one-stage training process of SNNs.
The number of epochs in fine-tuning phase for $i_{th}$ compression ratio in the budget list is set to the same value or scheduled as
$\frac{1}{S^i} \frac{T_{\text {epoch }}-C_{\text {epoch }}}{\sum_{j=i}^N \frac{1}{S^j}}$.
where, $S^i$ is the $i_{th}$ compression ratio. $T_{\text {epoch }}$ is the total number of epochs, $C_{\text {epoch }}$ is the number of already used epochs.
% \rev{cosine scheduler in cifar10net}

% As described, we train the compression and fine-tune jointly on SNN networks.

% The budget compression ratio in four methods has \emph{different} meaning. In both Flops and Lasso methods, it means the ratio of Flops. In the Uniform method, it means the ratio of Channels. In Latency constrained method, it is the ratio of latency values (inference time). We provide all compression training parameters and details in supplemental materials.

\begin{table}[!ht]
    \caption{Performance of our method and previous work IMP on CIFAR10-DVS dataset.}
    \small
    % \scriptsize
    \centering
        \begin{tabular}{c c c c c c}
            \toprule
            \textbf{Method} & \textbf{Arch.} & \textbf{Top-1 Acc. (\%)} & \textbf{Acc. (\%)} &  \textbf{Sparsity (\%)} \\
            \midrule
                \multirow{4}{*}{IMP~\cite{kim2022lottery}} & 
                \multirow{4}{*}{VGGSNN}  & 
                \multirow{4}{*}{82.80} &  81.10 & 76.20  \\ 
                & & & 81.50 & 86.57  \\ 
                & & & 80.10 & 89.91  \\
                & & & 78.60 & 94.28  \\
            \midrule
                \multirow{4}{*}{\textbf{Ours}} &
                \multirow{4}{*}{VGGSNN}  & 
                \multirow{4}{*}{82.80} &  \textbf{82.40} & \textbf{85.18}  \\ 
                & & & \textbf{81.90} & \textbf{90.14}  \\             
                & & & \textbf{81.20} & \textbf{93.14}  \\ 
                & & & \textbf{80.10} & \textbf{95.16} \\
                
            \bottomrule
            \end{tabular}
    \label{tab:cifar10dvs}
\end{table}
% \begin{table*}[!ht]
%     \caption{Performance of our method and previous work IMP on CIFAR10-DVS dataset.}
%     \centering
%         \resizebox{0.9\textwidth}{!}{
%         \begin{tabular}{c c c c c c}
%             \toprule
%             \textbf{Pruning Method} & \textbf{Dataset} & \textbf{Arch.} & \textbf{Baseline Acc. (\%)} & \textbf{Acc. (\%)} &  \textbf{Sparsity (\%)} \\
%             \midrule
%                 \multirow{4}{*}{IMP~\cite{kim2022lottery}} & 
%                 \multirow{4}{*}{CIFAR10-DVS} & 
%                 \multirow{4}{*}{VGGSNN}  & 
%                 \multirow{4}{*}{82.80} &  81.10 & 76.20  \\ 
%                 & & & & 81.50 & 86.57  \\ 
%                 & & & & 80.10 & 89.91  \\
%                 & & & & 78.60 & 94.28  \\
%             \midrule
%                 \multirow{4}{*}{\textbf{Ours}} & 
%                 \multirow{4}{*}{CIFAR10-DVS} & 
%                 \multirow{4}{*}{VGGSNN}  & 
%                 \multirow{4}{*}{82.80} &  \textbf{82.40} & \textbf{85.18}  \\ 
%                 & & & & \textbf{81.90} & \textbf{90.14}  \\             
%                 & & & & \textbf{81.20} & \textbf{93.14}  \\ 
%                 & & & & \textbf{80.10} & \textbf{95.16} \\
                
%             \bottomrule
%             \end{tabular}}
%     \label{tab:dvs}
% \end{table*}

\subsection{Quantitative Experiments}
\paragraph{Connection pruning and fine-tuning jointly.}
We use the connection of the SNNs model as the budget compression ratios for our Minimax optimization method. During compression training, the connection pruning and fine-tuning are trained jointly. Our joint compression method not only reduces the training time and simplifies the tedious fine-tuning process for different compression ratio, but also help the model under smaller ratios get better performance. Therefore, for one SNN model, we can achieve state-of-the-art (SOTA) performance of different ratios in one compression training process, which is different from previous work, since they only can get one compression ratio per training process~\cite{kim2022lottery}. 
As shown in Table~\ref{tab:first} and~\ref{tab:second}, we summed up the results of our compression experiments which were obtained by jointly pruning connection and fine-tuning the model.

\paragraph{Comparisons to previous work.}
We compare our method with previous works in Table~\ref{tab:first},~\ref{tab:second} and ~\ref{tab:cifar10dvs}.
The results of two shallow SNN models on MINST and CIFAR10 are shown in Table~\ref{tab:first}. We compare our connection pruning and fine-tuning jointly training method with previous research, including ADMM-based~\cite{admm2021}, Deep R~\cite{deepr2018}, Grad R~\cite{gradr2021}, and STDS~\cite{stdschen22}. It is observed that our approach outperforms previous pruning methods in terms of connection sparsity and accuracy on the two benchmark datasets. 
Furthermore, we have smaller $\Delta$Acc degradations under all comparable compression ratios.
 
We present the comparisons of deep SNN models, VGG16~\cite{vgg}, ResNet19~\cite{resnet} and SEW ResNet18~\cite{sewresnet} in Table~\ref{tab:second}. 
We reproduce the baseline based on the previous work~\cite{kim2022lottery}, and the performance is much higher than the reported result in~\cite{kim2022lottery}: 1.63\% higher on CIFAR10 and 3.37\% higher on CIFAR100 for ResNet19~\cite{resnet} SNN model, therefore, we compare absolute values of accuracy on deep SNNs. As shown in Table~\ref{tab:second}, our method is comparable to other approaches. For the VGG16 model, our methods have the highest accuracy at all connection sparsity ratios compared with Grad R~\cite{gradr2021} and IMP~\cite{kim2022lottery}. For ResNet19~\cite{resnet}, when the connection sparsity ratio is less than 97\%, our method significantly outperforms other methods. When the connection sparsity is higher than 97\%, we still achieve better accuracy compared with other works, but the connection sparsity is slightly smaller than Grad R~\cite{gradr2021} and IMP~\cite{kim2022lottery}. It is worth mentioning that the accuracy of our method can even be further improved compared to the baseline on all datasets when the sparsity is nearly 80\%. 
Even on the large-scale dataset like ImageNet, our pruning method has also achieved competitive performance compared with the state-of-the-art.

We also validated on neuromorphic datasets that have been less involved in previous
work. 
To the best of our knowledge, our work is the first work to compress SNNs on the temporal CIFAR10-DVS dataset.
Under the same structure and settings, we implemented the IMP method ~\cite{IMP2018lottery} and conducted our experiments on the CIFAR10-DVS dataset. As shown in Table~\ref{tab:cifar10dvs}, our method significantly improves the accuracy of VGGSNN~\cite{tettraining} model with different sparsity ratios.
In summary, experiments have shown that our compression optimization method can theoretically handle any type of SNN model.

\begin{table}[!ht]
\caption{Performance comparison of 6 Conv, 2 FC SNN model with different setting on CIFAR10 dataset.}
\centering
\resizebox{1.0\columnwidth}{!}{
\begin{tabular}{cccccc}
\toprule
\multicolumn{2}{c}{\begin{tabular}[c]{@{}c@{}}\textbf{a.Sequentially} \\ \textbf{w/o cos scheduler}\end{tabular}} &
  \multicolumn{2}{c}{\begin{tabular}[c]{@{}c@{}}\textbf{b.Jointly} \\ \textbf{w/o cos scheduler}\end{tabular}} &
  \multicolumn{2}{c}{\begin{tabular}[c]{@{}c@{}}\textbf{c.Jointly} \\ \textbf{w/ cos scheduler}\end{tabular}} \\ 
  \midrule
  Sparsity(\%) & Acc.(\%) & Sparsity(\%) & Acc.(\%) & Sparsity(\%) & Acc.(\%)\\ 
  \midrule 
  \multicolumn{1}{c}{\begin{tabular}[c]{@{}c@{}}88.04\\ 94.99\\ 97.67\\ 99.30\end{tabular}} &
  \begin{tabular}[c]{@{}c@{}}92.72\\ 91.95\\ 90.35\\ 80.9 \end{tabular} &
  \multicolumn{1}{c}{\begin{tabular}[c]{@{}c@{}}88.05\\ 95.01\\ 97.67\\ 99.36\end{tabular}} &
  \begin{tabular}[c]{@{}c@{}}92.61\\ 92.24\\ 91.47\\ 89.62 \end{tabular} &
  \multicolumn{1}{c}{\begin{tabular}[c]{@{}c@{}}88.04\\ 95.07\\ 97.71\\ 99.31\end{tabular}} &
  \begin{tabular}[c]{@{}c@{}}\textbf{93.40}\\ \textbf{93.29}\\ \textbf{92.75}\\ \textbf{90.32}\end{tabular} \\ 
 \bottomrule
\end{tabular}}
\label{tab: different_setting}
\end{table}

\section{Ablation Studies}
% \paragraph{Connection pruning and fine-tuning jointly}
% We use the connection of the SNNs model as the budget compression ratios. During compression training, connection pruning and fine-tuning are trained jointly. Our joint compression method, not only reduces the training time and simplifies the tedious fine-tuning process for different compression ratio, but also help smaller ratios get better performance. Therefore, for one SNN model, we can achieve state-of-the-art (SOTA) performance of different ratios in one compression training process. Unlike previous work, which only can get one compression ratio per training process. We summed up the results of our compression experiments which trained with connection pruning and fine-tuning jointly in Table \ref{tab:first} and Table \ref{tab:second}.

\paragraph{Comparison with sequential method}
We compare our end-to-end Minimax optimization method on both sequentially and jointly training,  for pruning and fine-tuning on the CIFAR10 dataset. For the sequential compression method, we first prune the SNNs models and save the pruned model snapshots during pruning training. We then fine-tune each of these snapshots for another 256 epochs for 6 Conv, 2 FC SNN models, which means need to fine-tune extra $n\times$ $256$ epochs, the $n$ is the length of the budget list. As shown in Table~\ref{tab: different_setting}, the compression ratios (column b) in which pruning and fine-tuning are trained jointly have better accuracy when compression ratios are smaller than 5\%. Sequentially trained compression ratios (column a) which connect 11.95\% have a slight advantage because trained more epochs than joint ratios with a similar connection. However, as the connection turn smaller, the joint method achieves better results and even obtains an 8.72\% accuracy advantage at a 0.64 connection ratio. What's more, the joint method in Table~\ref{tab: different_setting} was trained 700 epochs in total, which is far less than the train epochs numbers of the sequential method.

\paragraph{Comparison of fine-tuning options.}
We compare the final accuracy between applying with a cosine annealing scheduler and without any learning rate scheduler when fine-tuning on the CIFAR10 dataset. 
% Note, we take the 6 Conv, 2 FC snapshot with 75.00$\%$ sparsity in the normal compression as our initial model. 
For using a cosine annealing scheduler, in each stage of achieving the new compression ratio in the resource budget list, the number of the fine-tuning epoch is reset to 300 and the initial learning rate is changed to 0.001. As shown in Table~\ref{tab: different_setting}, the accuracy significantly increased when we use the cosine annealing scheduler in the fine-tuning phase.

% \section{Ablations}

% \section{Ablations}
% \paragraph{ Pruning and fine-tuning jointly}
% We use the connection of the SNNs model as the budget compression ratios. During compression training, the connection pruning and fine-tuning are trained jointly. Our joint compression method, not only reduces the training time and simplifies the tedious fine-tuning process for different compression ratio, but also help smaller ratios get better performance. Therefore, for one SNN model, we can achieve state-of-the-art (SOTA) performance of different ratios in one compression training process. Unlike previous work, which only can get one compression ratio per training process. We summed up the results of our compression experiments which trained with connection pruning and fine-tuning jointly in Table \ref{tab:first} and Table \ref{tab:second}.

\begin{table}[!htb]
    \caption{Model accuracy after pruning with different values of the learning rate $zlr$ on MNIST dataset.}
    \centering
    % \small
    \begin{tabular}{cccc}
        \toprule
        \multirow{2}{*}{\textbf{Sparsity(\%)}} & \multicolumn{3}{c}{\textbf{Accuracy(\%)}} \\ \cmidrule(l){2-4} & $zlr = 10^3$ & $zlr = 10^5$ & $zlr = 10^8$ \\
        \midrule
            75.00 & 98.85 & 98.79 & 98.80 \\
        \midrule
            85.00 & 98.75 & 98.69 & 98.66 \\ 
        \midrule
            95.00 & 97.68 & 97.39 & 97.16 \\
        \midrule
            97.00 & 96.21 & 95.97 & 95.60  \\ 
        \midrule
            98.70 & 91.57 & 91.40 & 90.71 \\
        \bottomrule
    \end{tabular}
    \label{tab:diffzlr}
\end{table}

\paragraph{Influence of the learning rate $zlr$.}
We evaluate the influence of the learning rate $zlr$ for dual variables $z$. We use gradient ascent to update $z$ as described in the Algorithm. Therefore, the value of the learning rate $zlr$ for dual variables $z$ influences the speed of pruning and the final performance. In Table~\ref{tab:diffzlr}, we show the accuracy of several compression ratios with different values of $zlr$ on the MNIST dataset. Because our compression method jointly optimizes the connectivity sparsity and model weights, a smaller $zlr$ makes the compression process slower and has a better balance between sparsity and performance. In Table~\ref{tab:diffzlr} we can see that the pruned model with smaller $zlr$ has higher accuracy.

\begin{figure}[t]
\centering
\includegraphics[width=0.9\linewidth]{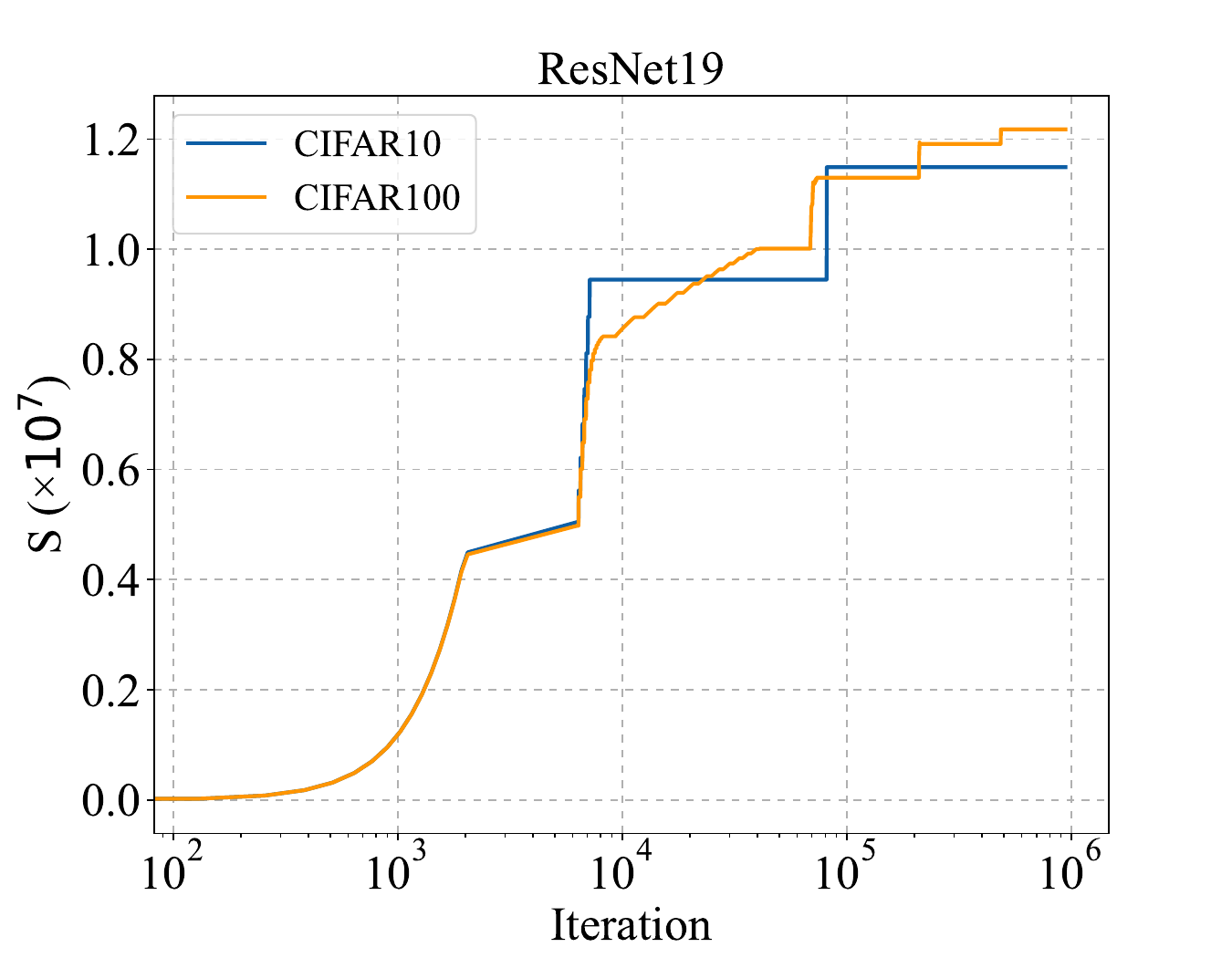}
\caption{The values of $s$ of ResNet19 on CIFAR10 and CIFAR100 datasets.}
\vspace{-5pt}
\label{fig:resnet19_s_data}
\end{figure}
\paragraph{Training trend plots}
We plot the changing trend of sparsity value $s$ in Figure~\ref{fig:resnet19_s_data}, which demonstrates that our Minimax optimization algorithm can converge well gradually.

\begin{figure}[!ht]
\centering
\includegraphics[width=0.45\textwidth]{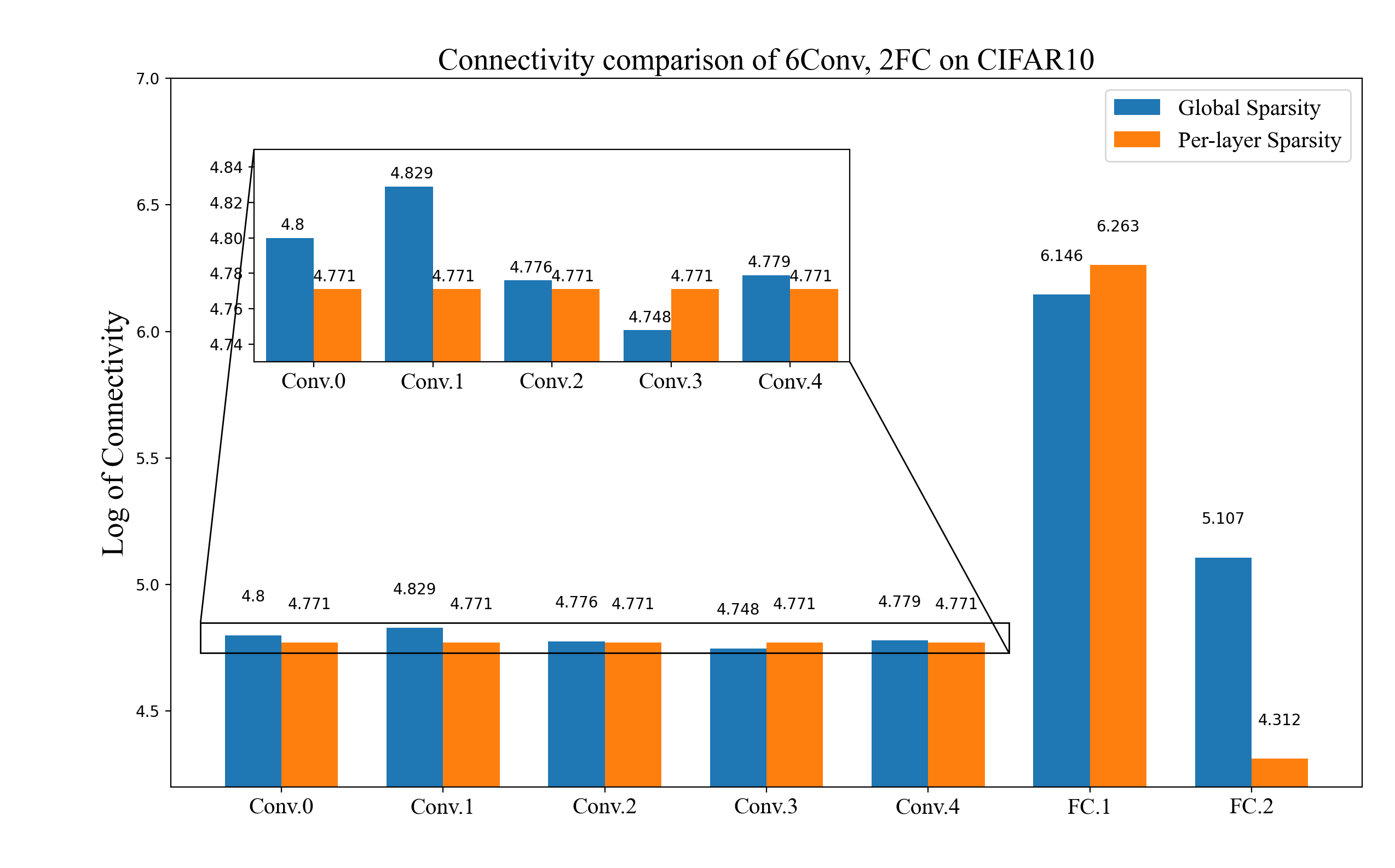}
\caption{Connectivity comparison of per-layer sparsity and global sparsity on CIFAR10.}
\label{fig:connenct_perlayer_global}
\end{figure}

\paragraph{Per-layer Sparsity vs. Global Sparsity}
We compare the difference between pruning using per-layer sparsity and our pruning using global sparsity. Our optimization is based on global sparsity $s$, which sorts the weight values of tensors of all layers together during pruning and coordinates the total sparsity of the whole model during optimization. On the contrary, the method using per-layer sparsity sorts the weights on each layer separately, each with a sparsity variable to control. 
For both the per-layer sparsity method and global sparsity method, our compression target is the layers whose connectivity is more than $1e4$.
Figure~\ref{fig:connenct_perlayer_global} shows the difference in connectivity sparsity between the per-layer sparsity method and our global sparsity method on 6 Conv, 2 FC SNN model, the 6 source convolution layers have the same connectivity in the baseline model. 
In Figure~\ref{fig:connenct_perlayer_global}, we can see the connectivity after pruning using the global sparsity method show more obvious diversities between these 6 convolution layers.
However, the per-layer sparsity method results in almost the same pruned connectivity for these 6 convolution layers.
In Table~\ref{tab:perlayer}, we can see that our global sparsity method has better pruning performance at each connectivity level.

\begin{table}[!ht]
\caption{Performance comparisons of per-layer sparsity control and global sparsity control on CIFAR10.}
% \small
\centering
\resizebox{\columnwidth}{!}{%
\begin{tabular}{@{}ccccc@{}}
\toprule
\textbf{Setting} &
  \multicolumn{2}{c}{\textbf{Per-layer}} &
  \multicolumn{2}{c}{\textbf{Global}} \\ \cmidrule(l){1-5} 
\multirow{4}{*}{\begin{tabular}[c]{@{}c@{}}CIFAR10\\ 6 Conv, 2 FC\end{tabular}} &
  Connectivity (\%) &
  Acc. (\%) &
  Connectivity (\%) &
  Acc. (\%) \\ \cmidrule(l){2-5} 
 &
  \begin{tabular}[c]{@{}c@{}}25.14\\ 12.93\\ 5.01\end{tabular} &
  \begin{tabular}[c]{@{}c@{}}91.47\\ 92.08\\ 92.00\end{tabular} &
  \begin{tabular}[c]{@{}c@{}}24.987\\ 11.95\\ 4.99\end{tabular} &
  \begin{tabular}[c]{@{}c@{}}\textbf{92.84}\\ \textbf{92.61}\\ \textbf{92.24}\end{tabular} \\ \bottomrule
\end{tabular}%
}
\label{tab:perlayer}
\end{table}

\section{Performance of structure pruning}
Aside from connection pruning, we extend our Minimax optimization method to structured pruning of the SNNs. Specifically, instead of using each weight value of tensors to participate in the sorting step in our algorithm, we use the reduced norm of each column from the tensor matrix to participate in the sorting. The weights of each column are updated simultaneously during the optimization.
In Table~\ref{tab:structureprune}, we show the performance of structured pruning on 6 Conv, 2 FC SNN model on the CIFAR10 dataset.

\begin{table}[!h]
    \caption{Performance of structured pruning on 6 Conv, 2 FC model.}
    \centering
    %\small
    \begin{tabular}{ccc}
        \toprule
        \multicolumn{2}{c}{\textbf{Compression Ratios}} & \multirow{2}{*}{\textbf{Acc. (\%)}} \\     
        \cmidrule(l){1-2} FLOPs & Parameters \\
        \midrule
            0.504 & 0.680 & 92.77 \\
        \midrule
            0.356 & 0.549 & 92.61 \\
        \midrule
            0.199 & 0.365 & 92.17 \\
        \midrule
            0.098 & 0.247 & 91.16\\
        \midrule
            0.051 & 0.163 & 89.23\\
        \bottomrule
        \end{tabular}
    \label{tab:structureprune}
\end{table}

\section{Conclusion}
In this paper, we present an end-to-end solution for the SNN compression method. We formulate the resource-constrained SNNs compression problem into a constrained optimization problem and jointly learn the connection sparsity and weights of SNN models. Our method effectively shrinks the model to meet the given budget list, which values can be connectivity, parameter size, FLOPs, and so on. Experiments demonstrate that our end-to-end Minimax optimization method can balance the performance and the computation efficiency of SNN models, and achieve state-of-the-art (SOTA) performance on different SNN models. In the future, we want to explore the efficiency and performance of SNN networks with neural architecture search (NAS) and other methods and consider the characteristics of the SNN models, such as SNN fire rate and time steps.
% \paragraph{Limitations}
% h-params 
% snn
% rate , 
% \paragraph{Future Work}
% structure
% nas

\bibliographystyle{ACM-Reference-Format}
\balance
\bibliography{refs}

\clearpage

\appendix

\section*{Appendix}

To better support the claims in the main text, we provide more related results and analysis in this supplementary material. 
We first present the surrogate gradient method used in the backward process during SNN training.
Then we analyze the effect of compression from scratch, instead of pre-trained models.
Thirdly, we show the memory and time cost for whole pruning phase on VGG16 and ResNet19.
Finally, we plot the performance comparisons to better visualize the superiority of our method.

\section{Surrogate-Gradient Based Learning}
\label{sec:backward_for_snn}
Unlike in ANNs, it is difficult to apply standard gradient based backpropagation in SNNs. Recently, methods based on surrogate gradient have provided an effective solution. To train deep SNNs, we use surrogate-gradient based Spatio-Temporal Backpropagation (STBP)~\cite{wu2018spatio}. 
With $L$ representing the loss function, the gradients $\partial L / \partial o_{i}^{t, n}$ and $\partial L / \partial u_{i}^{t, n}$ can be computed as follows:
\begin{align} 
\frac{\partial L}{\partial o_i^{t, n}} & =\sum_{j=1}^{l(n+1)} \frac{\partial L}{\partial o_i^{t, n+1}} \frac{\partial o_j^{t, n+1}}{\partial o_i^{t, n}}+\frac{\partial L}{\partial o_i^{t+1, n}} \frac{\partial o_i^{t+1, n}}{\partial o_i^{t, n}} 
\\ \frac{\partial L}{\partial u_i^{t, n}} & =\frac{\partial L}{\partial o_i^{t, n}} \frac{\partial o_i^{t, n}}{\partial u_i^{t, n}}+\frac{\partial L}{\partial o_i^{t+1, n}} \frac{\partial o_i^{t+1, n}}{\partial u_i^{t, n}}
\end{align}
Where $t$ denotes the time step $t$, $n$ denotes the $nth$ layer and 
$l(n)$ denotes the number of neurons in the $nth$ layer.
$o_i$ and $u_i$ are the output and neuronal potential of the $ith$ neuron, respectively.
Due to the non-differentiable property of the binary spike activities, $\partial o_k / \partial u_k$ cannot be derived. We utilize shifted ArcTan function $h(u)$ to approximate the derivative of spike activity following previous works~\cite{chen2021pruning, fang2021deep, deng2022temporal}, which is defined by
\begin{align} 
    h(u)=\frac{1}{\pi} \arctan (\pi u)+\frac{1}{2}
\end{align}

\section{Effect of Compression from Scratch}
\label{sec:compression_from_scratch}
A key component in successful extreme compression of SNNs is proper initialization of weights. 
As mentioned in the main manuscript, we use the pre-trained SNN model for compression, which contains redundant parameters and structures.
Another approach is pruning from randomly initialized weights.  
Here, we select shallow network 2 FC and deep network VGG16 to show the effect of compression from pre-trained model and scratch in Table~\ref{tab:scratch}. 
As can be seen, the performance on compression from scratch is worse than compression from pre-trained.
\begin{table}[!hb]
    \caption{Compression performance from pre-training and scratch.}
    \centering
        \begin{tabular}{c c c c c}
            \toprule
            \multirow{2}{*}{\textbf{Arch.}} & \multirow{2}{*}{\textbf{Dataset}} & \multicolumn{2}{c}{\textbf{Acc. (\%)}} &  \multirow{2}{*}{\begin{tabular}[c]{@{}c@{}}\textbf{Sparsity}\\ \textbf{(\%)} \end{tabular}} \\
             \cmidrule(l){3-4} & & \textbf{Pre-trained} & \textbf{Scratch}\\ 
            \midrule
                \multirow{5}{*}{2 FC}  &  \multirow{5}{*}{MNIST} &  \textbf{98.85} & 98.35 & 75.00  \\
                & & \textbf{98.75} & 97.93 & 85.00  \\ 
                & & \textbf{97.68} & 96.92 & 95.00  \\
                & & \textbf{96.21} & 96.09 & 97.00  \\
                & & \textbf{91.57} & 91.35 & 98.70  \\
            \midrule
                \multirow{3}{*}{VGG16}  &  \multirow{3}{*}{CIFAR10} &  \textbf{93.26} & 90.78 & 90.09  \\
                & & \textbf{92.76} & 89.89 & 95.70  \\ 
                & & \textbf{92.00} & 90.37 & 98.13  \\
            \bottomrule
            \end{tabular}
    \label{tab:scratch}
\end{table}

\section{Memory Usage and Training Time}
To provide the reference on resource consumption and compression time measured in the main manuscript, we take VGG16 and ResNet19 as examples to provide our method's GPU memory usage and total training time (feedforward + backward) in Table~\ref{tab:memory_time}. We use SpikingJelly framework which implements optimized SNN neuron and the training is conducted on one Nvidia V100 GPU with batch size 32. As we can see, ResNet19 requires more execution time and GPU memory than VGG16.
\begin{table}[!b]
    \caption{Resource consumption and training time of our method.}
        \begin{tabular}{c c c c}
            \toprule
            \textbf{Arch.} & \textbf{Dataset} & \textbf{Time (d)} &  \textbf{Memory (M)} \\
            \midrule
                 VGG16 &  CIFAR10 &  0.5 & 2452  \\ 
                 VGG16 &  CIFAR100 & 0.7 & 2454  \\ 
                 ResNet19 & CIFAR10 & 3.8 & 6502  \\ 
                 ResNet19 & CIFAR100 & 4.5 & 6506  \\      
            \bottomrule
            \end{tabular}
    \label{tab:memory_time}
\end{table}

\section{More Results on Neuromorphic Datasets}
We have shown the results on the DVS-CIFAR10 dataset in Table 3 of our original manuscript, in which we have achieved SOTA performance. Here, we perform more comparative experiments on different neuromorphic datasets such as DVS128 Gesture and N-Caltech 101. For DVS128 Gesture, the network architecture is (2 Conv, 2 FC) MP4-64C3-LIF-AP2-128C3-LIF-AP2-0.5DP-300FC-LIF-11FC-LIF. The time step is 20 and the batch size is 32. For N-Caltech 101, the network architecture is (4 Conv, 2 FC) 32C3-LIF-MP2-64C3-LIF-MP2-128C3-LIF-MP2-128C3-LIF-MP4-0.8DP-1024FC-LIF-101FC-LIF. The time step is 14 and the batch size is 16.
Where $x$C$y$ denotes the Conv2D layer with output channels = $x$ and kernel size = $y$,
MP$y$ denotes the MaxPooling layer with kernel size = $y$,
AP$y$ denotes the AvgPooling layer with kernel size = $y$,
$n$FC denotes the Fully Connected layer with output feature = $n$,
$m$DP is the spiking DroPout layer with dropout ratio $m$.
For both of architectures, the learning rate is 0.001, the loss function is MSE and the optimizer is Adam.

In order to compare with other SOTA SNN compression methods, we apply the SOTA method IMP~\cite{IMP2018lottery} and STDS~\cite{stdschen22} on the above two network architectures with these two datasets respectively.
As shown in the Table~\ref{tab:neuromorphic_datasets}, it is evident that our method outperforms existing SOTA approaches on different neuromorphic datasets at different levels of sparsities by a large margin. This also means that our optimization framework performs very well in handling the data with spatial features, as well as the data with spatiotemporal features.

\begin{table*}
    \caption{Performance comparison between our method and previous works on DVS128 Gesture and N-Caltech 101 datasets.}
    \centering

    \begin{tabular}{c c c c c c}
        \toprule
        \textbf{Pruning Method} & \textbf{Dataset} & \textbf{Arch.} & \textbf{Top-1 Acc. (\%)}& \textbf{Acc. (\%)} & \textbf{Sparsity (\%)} \\
        \midrule
            \multirow{5}{*}{IMP~\cite{IMP2018lottery}} & 
            \multirow{5}{*}{DVS128 Gesture} & 
            \multirow{5}{*}{2 Conv, 2 FC}  & 
            \multirow{5}{*}{93.75} & 92.71 & 75  \\ 
            & & & & 92.71 & 85  \\ 
            & & & & 92.36 & 90  \\
            & & & & 91.32 & 93  \\
            & & & & 89.58 & 95  \\
        \midrule
            \multirow{5}{*}{STDS~\cite{stdschen22}} & 
            \multirow{5}{*}{DVS128 Gesture} & 
            \multirow{5}{*}{2 Conv, 2 FC}  & 
            \multirow{5}{*}{93.75} & 90.28 & 75  \\ 
            & & & & 88.19 & 85  \\ 
            & & & & 89.58 & 90  \\
            & & & & 91.32 & 93  \\
            & & & & 88.54 & 95  \\
        \midrule
            \multirow{5}{*}{Ours} & 
            \multirow{5}{*}{DVS128 Gesture} & 
            \multirow{5}{*}{2 Conv, 2 FC}  & 
            \multirow{5}{*}{93.75} & \textbf{93.75} & \textbf{75}  \\ 
            & & & & \textbf{93.40} & \textbf{85}  \\ 
            & & & & \textbf{93.06} & \textbf{90}  \\
            & & & & \textbf{93.06} & \textbf{93}  \\
            & & & & \textbf{92.71} & \textbf{95}  \\  
        \midrule
        \midrule
            \multirow{5}{*}{IMP~\cite{IMP2018lottery}} & 
            \multirow{5}{*}{N-Caltech 101} & 
            \multirow{5}{*}{4 Conv, 2 FC}  & 
            \multirow{5}{*}{78.86} & 75.21 & 75  \\ 
            & & & & 68.65 & 85  \\ 
            & & & & 63.06 & 90  \\
            & & & & 58.69 & 93  \\
            & & & & 50.30 & 95  \\
        \midrule
            \multirow{5}{*}{STDS~\cite{stdschen22}} & 
            \multirow{5}{*}{N-Caltech 101} & 
            \multirow{5}{*}{4 Conv, 2 FC}  & 
            \multirow{5}{*}{78.86} & 74.32 & 75  \\ 
            & & & & 74.97 & 85  \\ 
            & & & & 72.78 & 90  \\
            & & & & 69.14 & 93  \\
            & & & & 66.59 & 95  \\
        \midrule
            \multirow{5}{*}{Ours} & 
            \multirow{5}{*}{N-Caltech 101} & 
            \multirow{5}{*}{4 Conv, 2 FC}  & 
            \multirow{5}{*}{78.86} & \textbf{76.55} & \textbf{75}  \\ 
            & & & & \textbf{75.58} & \textbf{85}  \\ 
            & & & & \textbf{75.33} & \textbf{90}  \\
            & & & & \textbf{74.61} & \textbf{93}  \\
            & & & & \textbf{68.89} & \textbf{95}  \\  
        \bottomrule
        \end{tabular}
    \label{tab:neuromorphic_datasets}
\end{table*}

\section{Visualization of Performance Comparisons}
\begin{figure*}[!h]
\centering
\begin{subfigure}{0.49\linewidth}
\includegraphics[width=\linewidth]{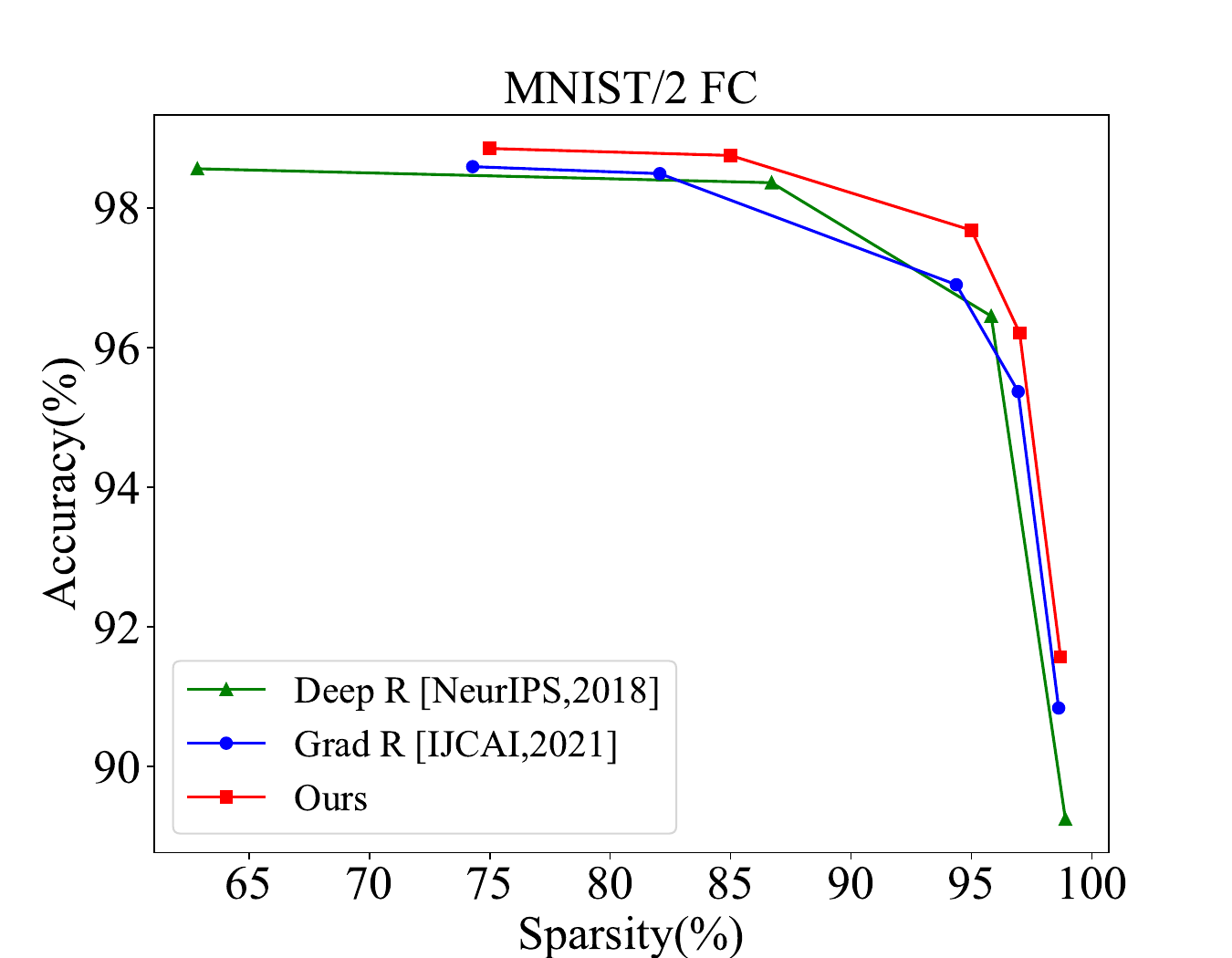}
\caption{Acc. on MNIST}
\end{subfigure}
\begin{subfigure}{0.49\linewidth}
\includegraphics[width=\linewidth]{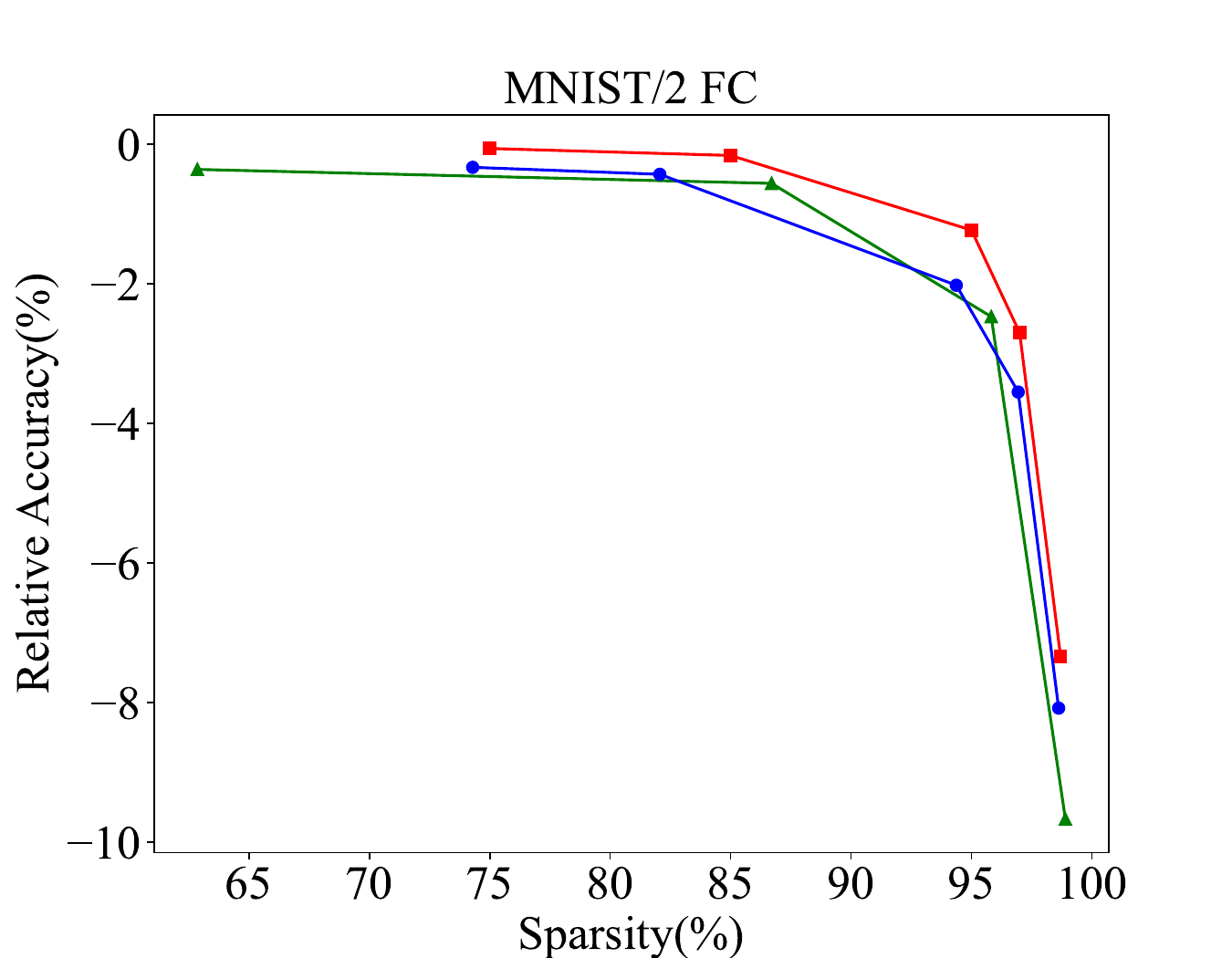}
\caption{Relative Acc. on MNIST}
\end{subfigure}
\caption{Performance comparisons between our method and previous work on 2 FC.}
\label{fig:cifar10net_results_2fc}
\end{figure*}

\begin{figure*}[!h]
\centering
\begin{subfigure}{0.49\linewidth}
\includegraphics[width=\linewidth]{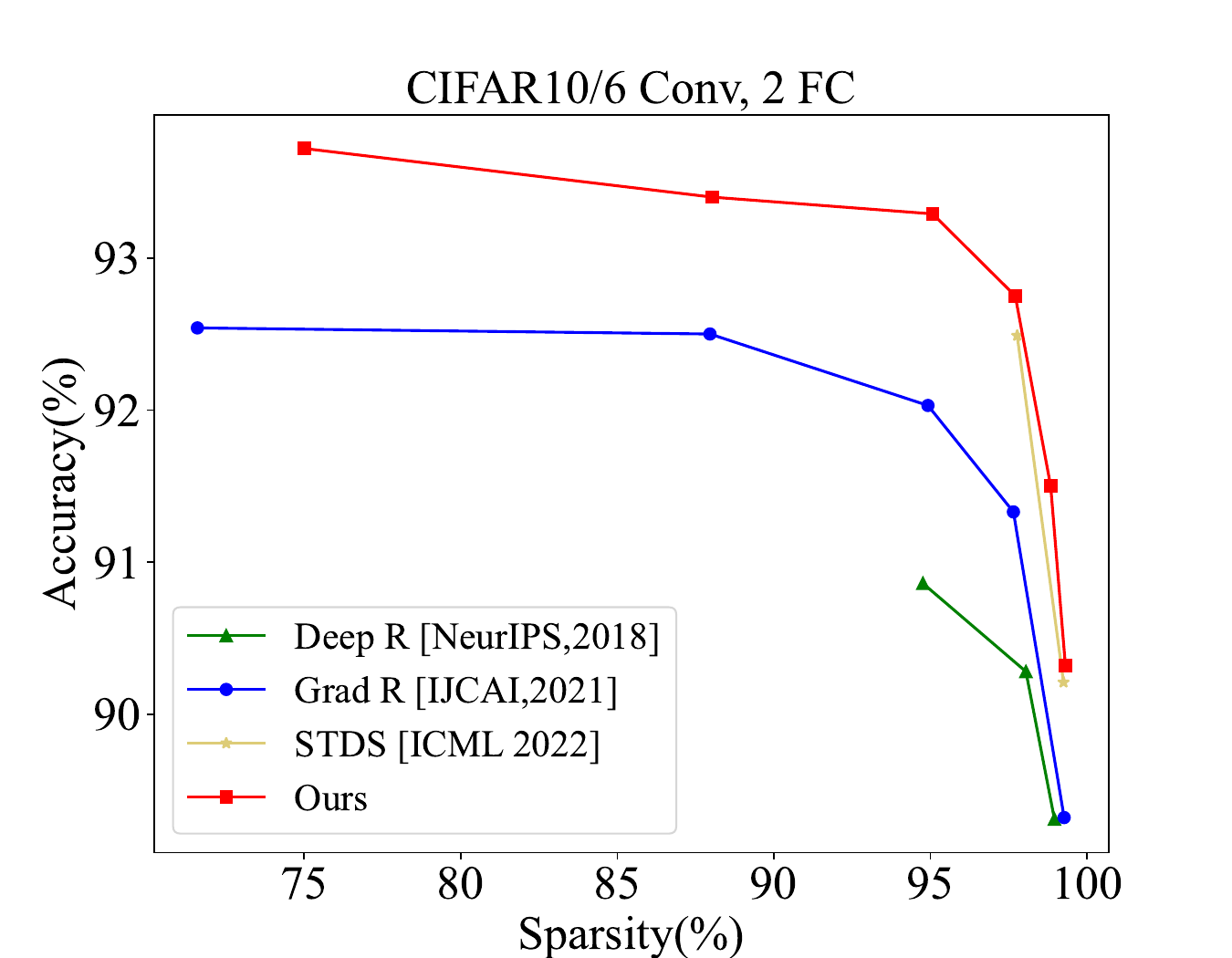}
\caption{Acc. on CIFAR10}
\end{subfigure}
\begin{subfigure}{0.49\linewidth}
\includegraphics[width=\linewidth]{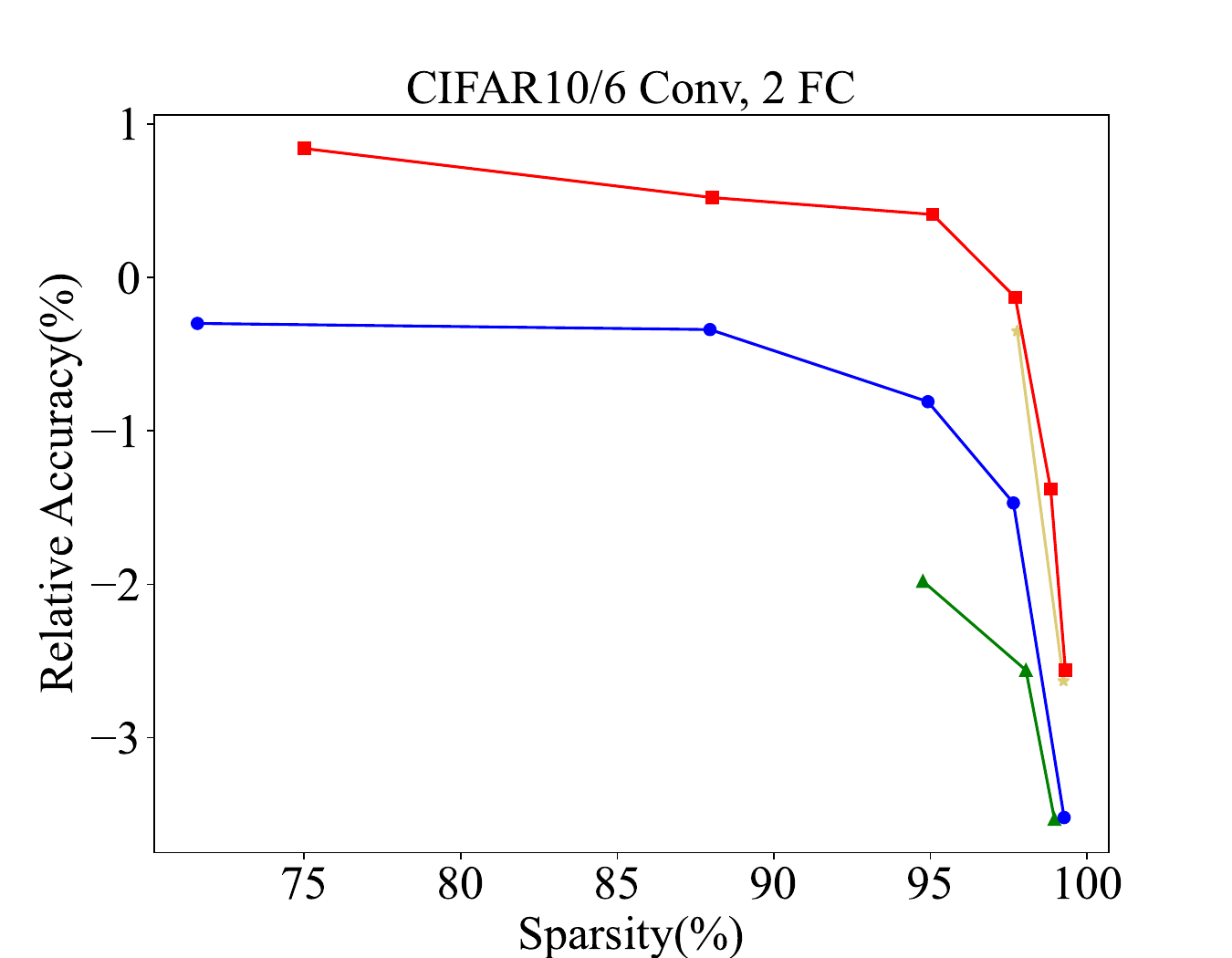}
\caption{Relative Acc. on CIFAR10}
\end{subfigure}
\caption{Performance comparisons between our method and previous work on 6 Conv, 2 FC.}
\label{fig:cifar10net_results_6conv2fc}
\end{figure*}

\begin{figure*}[!h]
\centering
\begin{subfigure}{0.49\linewidth}
\includegraphics[width=\linewidth]{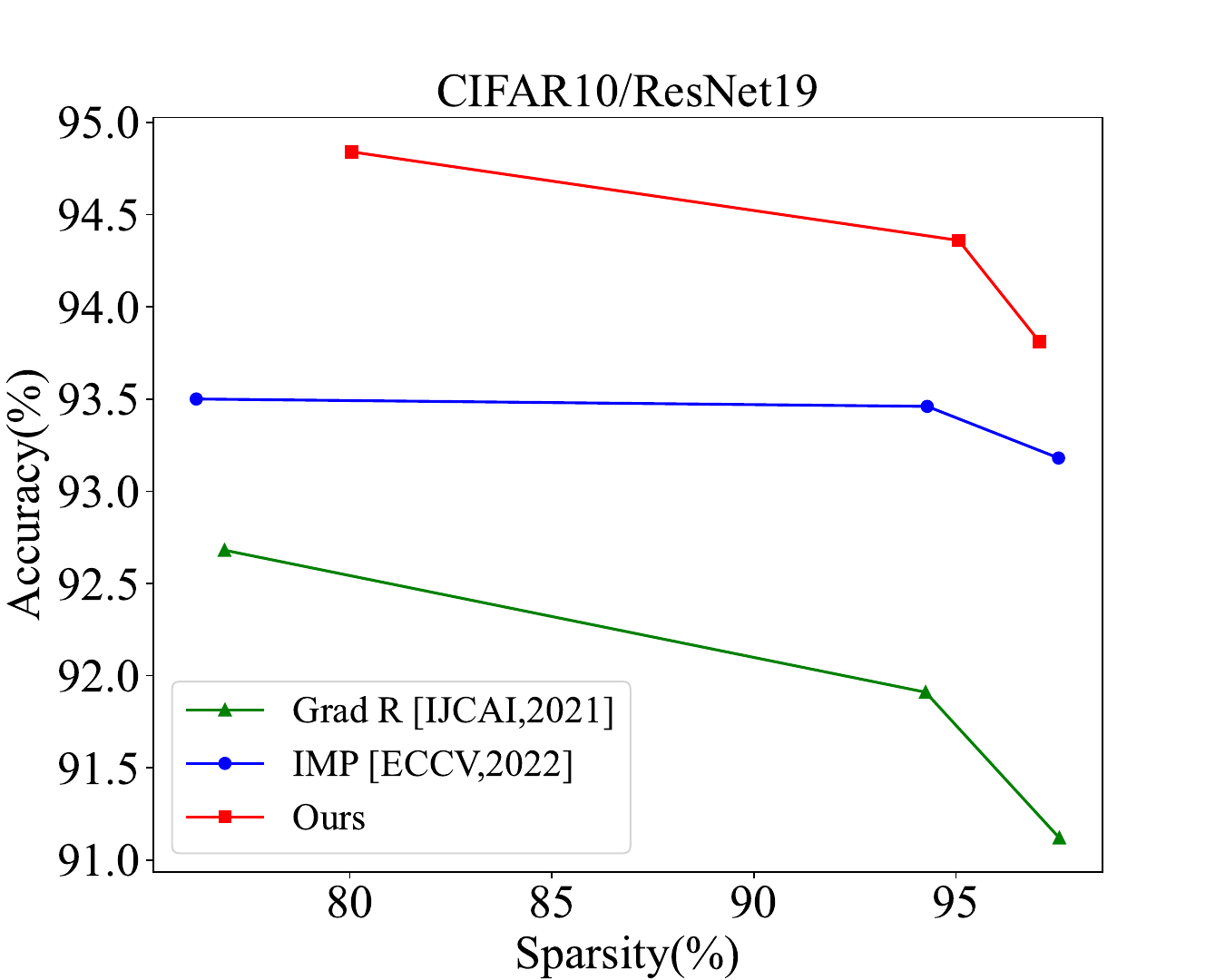}
\end{subfigure}
\begin{subfigure}{0.49\linewidth}
\includegraphics[width=\linewidth]{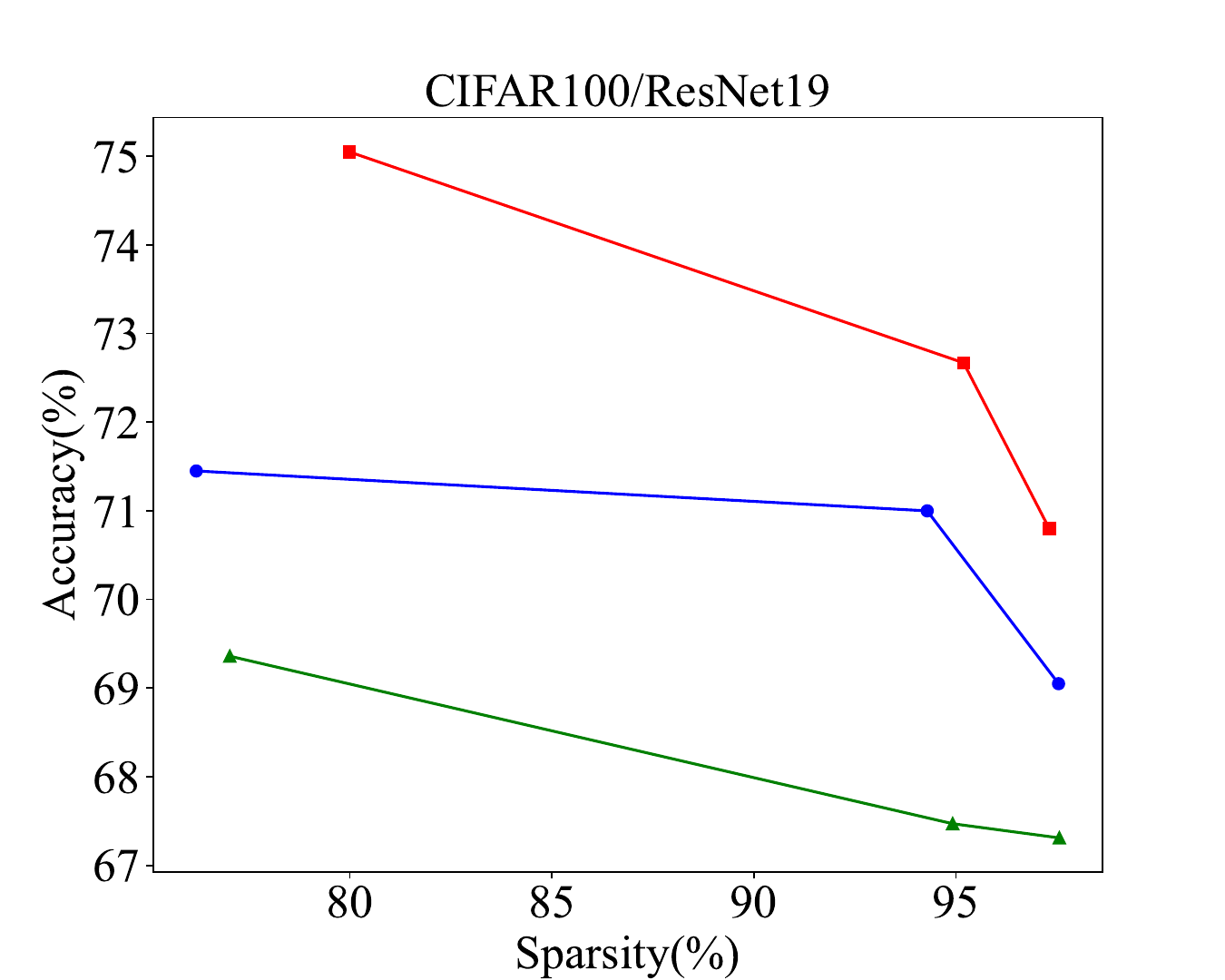}
\end{subfigure}
\caption{Performance comparisons between our method and previous work on ResNet19.}
\label{fig:res19}
\end{figure*}

\begin{figure*}[!h]
\centering
\begin{subfigure}{0.49\linewidth}
\includegraphics[width=\linewidth]{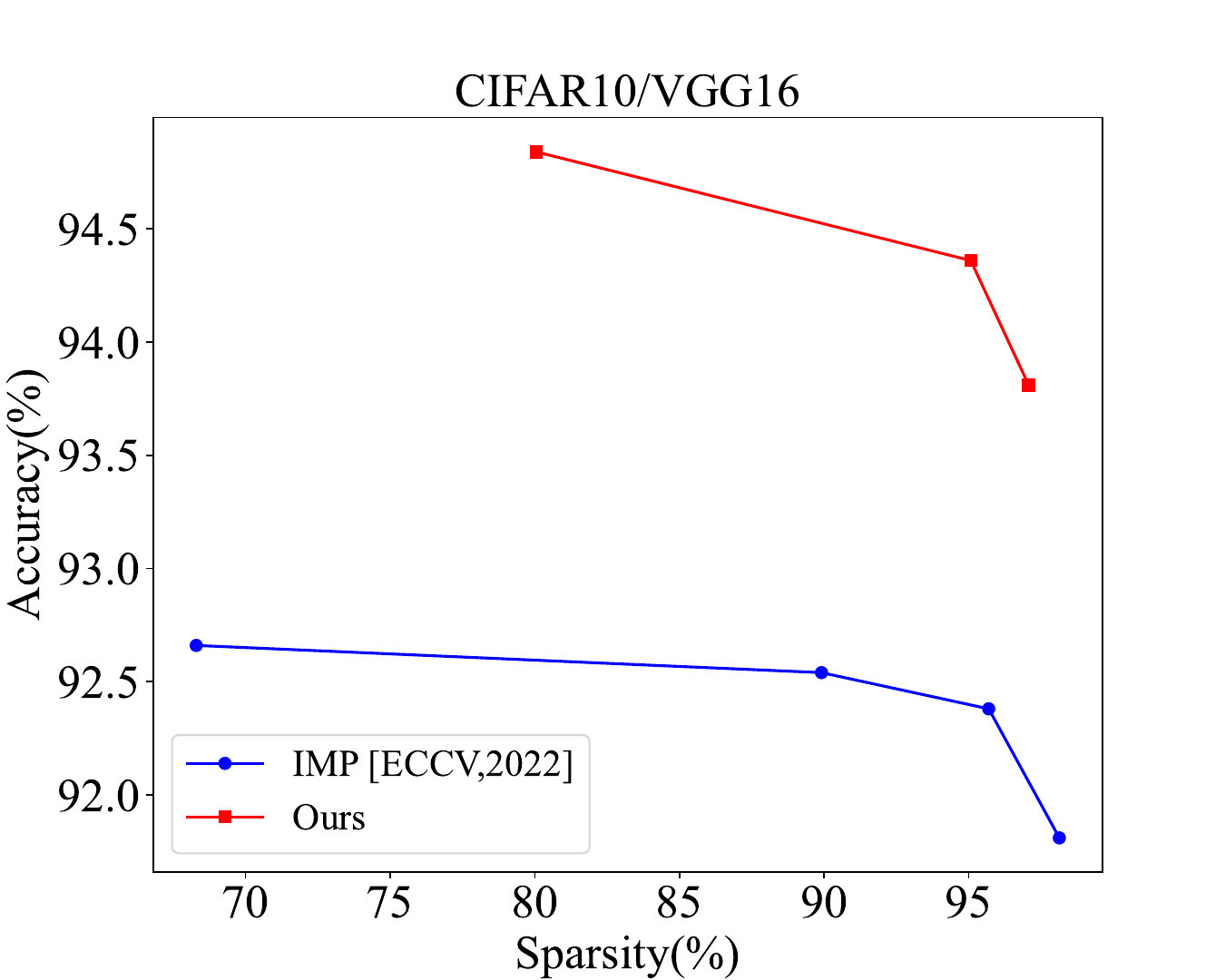}
\end{subfigure}
\begin{subfigure}{0.49\linewidth}
\includegraphics[width=\linewidth]{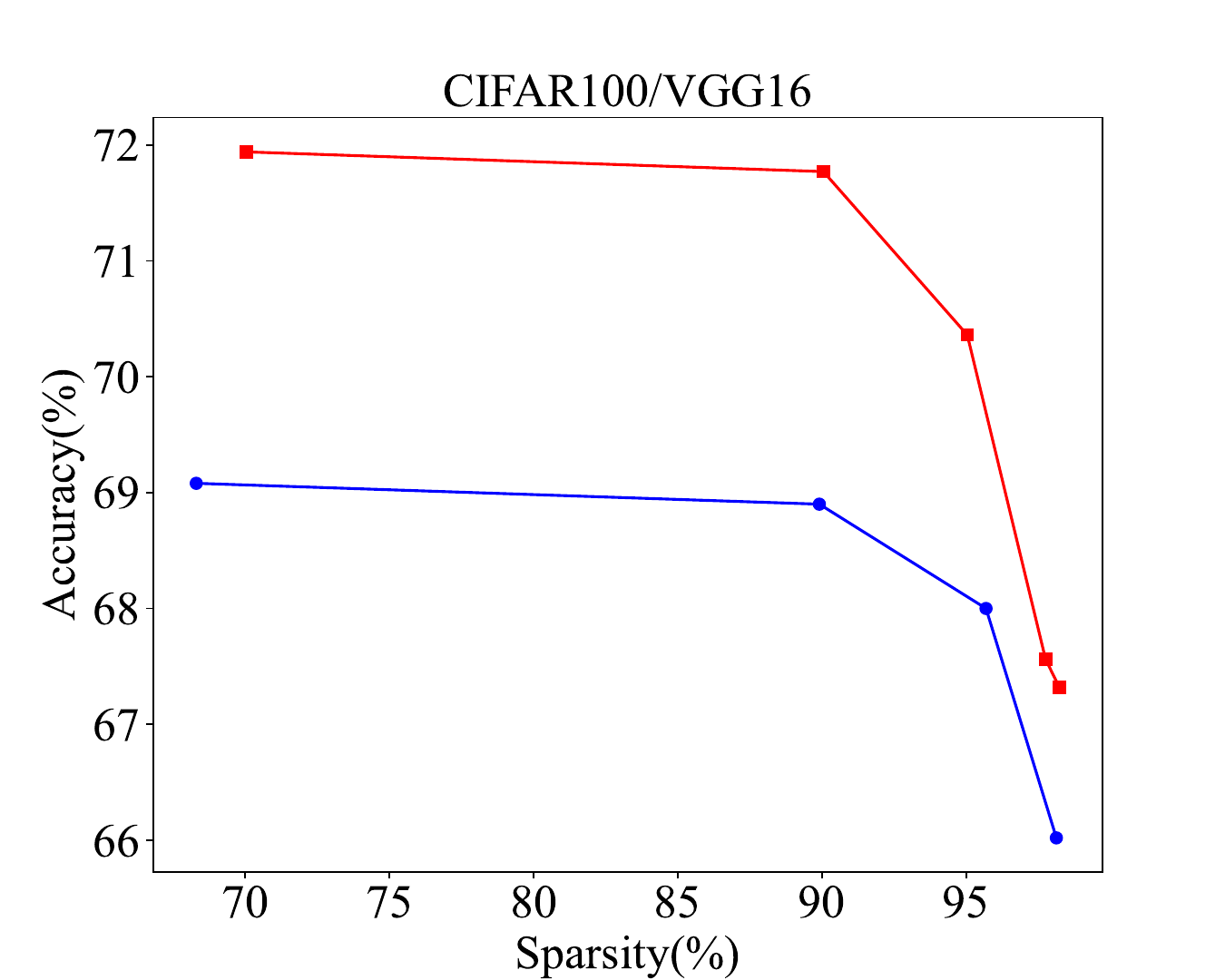}
\end{subfigure}
\caption{Performance comparisons between our method and previous work on VGG16.}
\label{fig:vgg16}
\end{figure*}

For SNN models on MNIST, CIFAR10 and CIFAR100 datasets, we plot the figure of absolute accuracy versus sparsity and the figure of relative accuracy drop versus sparsity. Since previous methods did not have exact sparsity levels with ours. \textbf{The plots can better visualize our method's superiority.} As shown in Figure~\ref{fig:cifar10net_results_2fc}, Figure~\ref{fig:cifar10net_results_6conv2fc}, Figure~\ref{fig:res19}, and Figure~\ref{fig:vgg16}, our method outperforms other methods by a clear margin.

\end{document}